\documentclass[lettersize,journal]{IEEEtran}

\usepackage{amsmath,amsfonts}
\usepackage{algorithmic}
\usepackage{array}
\usepackage[caption=false,font=normalsize,labelfont=sf,textfont=sf]{subfig}
\usepackage{textcomp}
\usepackage{stfloats}
\usepackage{url}
\usepackage{verbatim}
\usepackage{graphicx}
\def\BibTeX{{\rm B\kern-.05em{\sc i\kern-.025em b}\kern-.08em
    T\kern-.1667em\lower.7ex\hbox{E}\kern-.125emX}}
\usepackage{balance}

\usepackage{enumitem}
\usepackage{makecell}
\usepackage{booktabs}

\newcommand{\problem}[0]{performance-reproducibility problem}

\newcommand{\Problemone}[0]{UQD-Problem $1$}
\newcommand{\Problemtwo}[0]{UQD-Problem $2$}
\newcommand{\Problemthree}[0]{UQD-Problem $3$}

\newcommand{\parametrisation}[0]{$\delta$-parametrisation}
\newcommand{\paramf}[0]{\delta_f}
\newcommand{\paramr}[0]{\delta_r}

\usepackage{booktabs} 
\usepackage{makecell}

\title{
    Exploring the Performance-Reproducibility Trade-off in Quality-Diversity
}

\author{Manon Flageat, Hannah Janmohamed, Bryan Lim and Antoine Cully \\ Adaptive and Intelligent Robotics Lab, Department of Computing, Imperial College London, UK}

\begin{document}

\maketitle

\begin{abstract}
Quality-Diversity (QD) algorithms have exhibited promising results across many domains and applications. However, uncertainty in fitness and behaviour estimations of solutions remains a major challenge when QD is used in complex real-world applications. While several approaches have been proposed to improve the performance in uncertain applications, many fail to address a key challenge: determining how to prioritise solutions that perform consistently under uncertainty, in other words, solutions that are reproducible. Most prior methods improve fitness and reproducibility jointly, ignoring the possibility that they could be contradictory objectives. For example, in robotics, solutions may reliably walk at 90\% of the maximum velocity in uncertain environments, while solutions that walk faster are also more prone to falling over. 
As this is a trade-off, neither one of these two solutions is ``better" than the other. 
Thus, algorithms cannot intrinsically select one solution over the other, but can only enforce given preferences over these two contradictory objectives.
In this paper, we formalise this problem as the performance-reproducibility trade-off for uncertain QD. 
We propose two new a-priori QD algorithms that find efficient solutions for given preferences over the trade-offs. We also propose an a-posteriori QD algorithm for when these preferences cannot be defined in advance. Our results show that our approaches successfully find solutions that satisfy given preferences. Importantly, by simply accounting for this trade-off, our approaches perform better than existing uncertain QD methods.
This suggests that considering the performance-reproducibility trade-off unlocks important stepping stones that are usually missed when only performance is optimised.
\end{abstract}


\section{Introduction}

\IEEEPARstart{M}{any} problems, such as decision-making and content generation, benefit greatly from a variety of solutions and options. 
This provides a range of choices of high-performing solutions to the final user~\cite{map_elites} or constitutes reservoirs of alternative solutions in unexpected situations~\cite{nature}.
Quality-Diversity (QD) algorithms~\cite{pugh, book_chapter, framework} have attracted attention in recent years as a method to find such collections of diverse and high-performing solutions.
QD has been applied to a range of problems: finding diverse gaits~\cite{nature} or grasps~\cite{huber2023quality} in robotics, creating innovative designs~\cite{design, janmohamed2024multi}, or generating content for video games~\cite{video_games, video_games_2}.
While uncovering diversity remains the primary usage of these algorithms, they have also proven promising in helping to solve deceptive tasks~\cite{ns, chalumeau2022neuroevolution}, as they discover stepping stones toward more promising solutions that can hardly be found with directed search~\cite{stepping_stones}.

\begin{figure}[t]
  \centering
  \includegraphics[width=0.95\hsize]{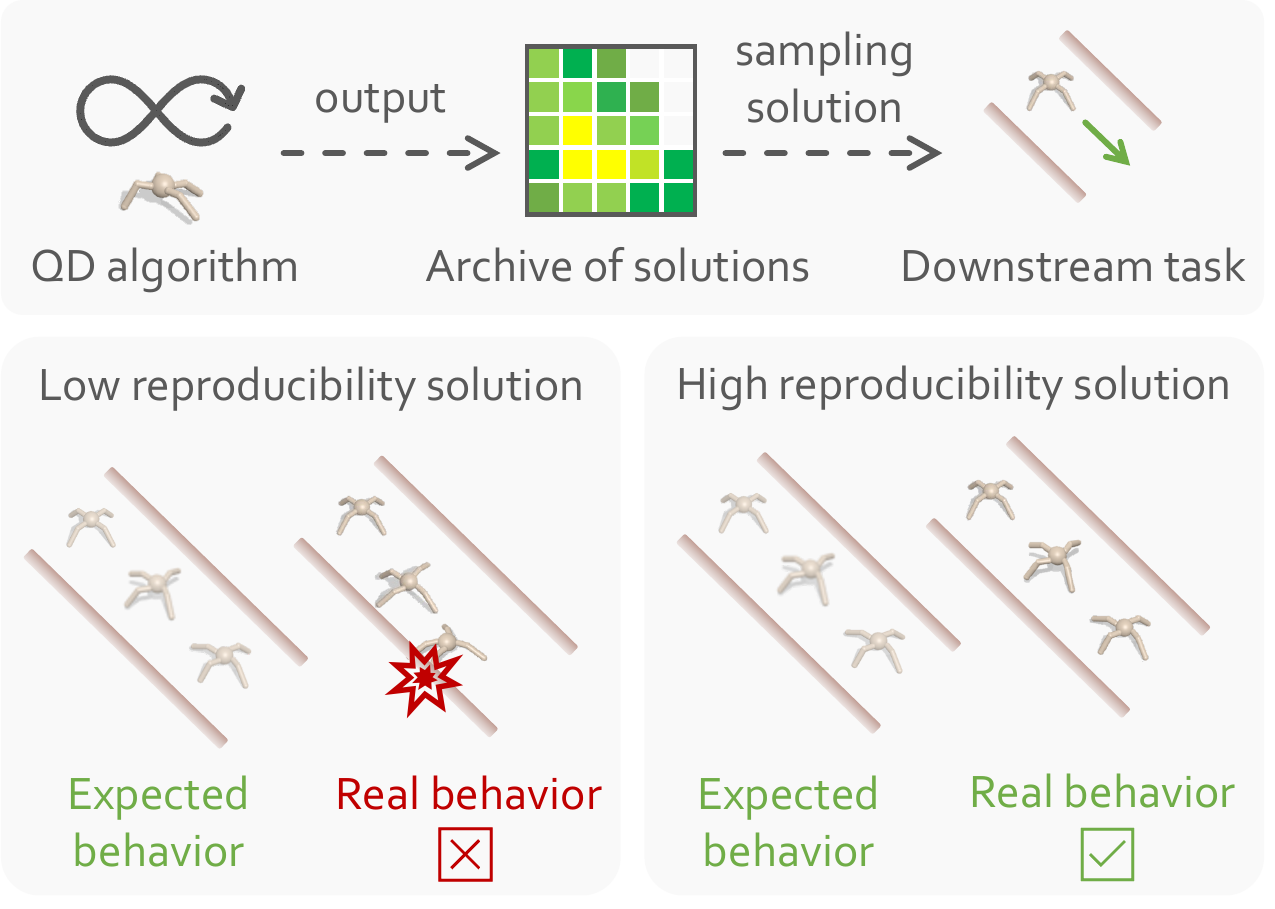}
  \caption{
  \textbf{Importance of reproducibility:}
  A QD algorithm produces an archive of solutions that is then used to solve a downstream task (top). 
  Solutions with low reproducibility fail to reproduce their behaviour when deployed thus failing to solve the downstream task (bottom).
  }
  \label{fig:reproducibility}
\end{figure}

However, in uncertain domains, where environment dynamics are stochastic or sensors are noisy, QD algorithms struggle to fulfil their objectives~\cite{flageat2023uncertain, hbr, glette_stochastic}. 
Due to their inherent elitism, they tend to favour solutions that are lucky and appear more promising than they truly are~\cite{ flageat2023uncertain}. 
This setting, where vanilla QD approaches are limited due to uncertainty, is referred to as Uncertain QD (UQD)~\cite{flageat2023uncertain}.
New QD algorithms have been proposed to handle the UQD setting~\cite{flageat2023uncertain, flageat2020fast, mace2023quality, adaptive, grillotti2023don}.
Some of these methods have highlighted the importance of not only dealing with lucky solutions but also prioritising solutions that perform consistently under the uncertainty distribution. 
For example, consider the robotic locomotion task illustrated in Figure~\ref{fig:reproducibility}, it is undesirable to find a solution that runs fast if it only manages to run $70\%$  of the time but falls or collides with obstacles the remaining $30\%$. 
Instead, when finding a set of solutions that can move to different target positions, we want solutions that consistently reach their target position all or most of the time. 
This consistency in the behaviour is referred to as reproducibility~\cite{flageat2023uncertain}. 
Ensuring reproducibility of solutions is critical for the use of QD in real-world applications (Figure~\ref{fig:reproducibility}). 

Prior work in UQD has treated optimising performance and reproducibility as a single goal~\cite{flageat2023uncertain, grillotti2023don, mace2023quality}.
However, these two objectives may not be aligned. 
For example, in robotic control environments, fast-moving solutions may be intrinsically more brittle than more conservative, slower solutions. 
Prior work takes the approach of first finding solutions that are high-performing and then, if possible, make these solutions more reproducible~\cite{flageat2023uncertain, grillotti2023don, mace2023quality}.
Implicitly, these algorithms assume that there are solutions that are both high-performing and reproducible. 
On the contrary, we argue that it could be impossible to find solutions that are high-performing and reproducible across every niche. 
Thus, there exists a trade-off between performance and reproducibility of solutions that has been largely underappreciated in previous work.

A key consideration of the performance-reproducibility trade-off is that there is no ``best" solution.
For example, in a robotic control task, one solution could achieve $100\%$ maximum velocity, but a reproducibility score of $80\%$ while another achieves $80\%$ maximum velocity and $100\%$ reproducibility score.
The ``better" solution depends on the user's preference between performance and reproducibility.
Hence, we argue that UQD algorithms cannot ``solve" the performance-reproducibility trade-off problem. 
They should instead aim to find the trade-off which satisfies the performance vs. reproducibility preferences specified by a user or a task.
If such a preference is unknown, they should return a variety of solutions which provide different trade-offs, such that the user can choose retrospectively.

The UQD literature usually considers two main problems~\cite{flageat2023uncertain}: the performance estimation problem (i.e. how to best estimate the true quality and diversity of a solution), that we label \Problemone{}, and the reproducibility problem (i.e. how to prefer reproducible solutions), that we label \Problemtwo{}. 
We propose an additional, third problem: the \textbf{\problem{}}, \Problemthree{}, which focuses on enforcing a specified performance-reproducibility trade-off.
Our contributions are as follows:
\begin{itemize}
    \item We propose a systematic way to formulate performance-reproducibility preferences: the \parametrisation{}.
    \item We develop $2$ new UQD approaches when preferences are available \textit{a-priori}. 
    Interestingly, one of them constitutes the general case of the previously proposed ME-LS~\cite{mace2023quality}.
    \item We propose a multi-objective UQD approach that aims to return a set of Pareto-optimal trade-offs in each niche when preferences are only available \textit{a-posteriori}. 
    \item We extend the existing UQD Benchmark~\cite{flageat2023benchmark} with $4$ new tasks for the \problem{}.
    \item Finally, we conduct a rigorous experimental study across $6$ benchmark tasks and $3$ robotic tasks used in previous UQD works and against a wide variety of baselines.
\end{itemize}
Our results show that our proposed approaches succeed in solving the \problem{} in UQD.
However, despite not being designed to produce better archives than existing UQD methods, our results indicate that our approaches perform generally better (i.e. higher QD-Score). 
Specifically, our approaches, which were only designed to address \Problemthree{} (enforcing preferred performance-reproducibility trade-offs), outperform UQD baselines on \Problemone{} (performance-estimation) and \Problemtwo{} (reproducibility-maximisation). 
This result indicates that accounting for the performance-reproducibility trade-off is key to future QD approaches in uncertain domains.

\section{Background and Related Work}

\subsection{Quality-Diversity} \label{sec:qd}

Quality-Diversity algorithms (QD)~\cite{book_chapter, pugh, framework} find collections of diverse and high-performing solutions to an optimisation problem. 
A QD algorithm thus returns a collection (or archive) $\mathcal{A}$ of solutions that are all as diverse as possible according to some dimensions of diversity, and as high-performing as possible according to a fitness function. 
These dimensions of diversity are referred to as features (also called behaviour descriptors~\cite{map_elites}, or measures~\cite{fontaine2020covariance} in the literature) and they are usually defined as part of the task.

\subsubsection{MAP-Elites}

The most common QD approach is MAP-Elites (ME)~\cite{map_elites}. 
ME discretises the feature space into a grid of cells, and keeps only the highest performing solution, an elite, in each cell. 
This grid of solutions is returned by the algorithm as the final collection $\mathcal{A}$. 
To find the elites, ME improves $\mathcal{A}$ iteratively across a fixed number of generations $N$. At each generation, a batch of solutions is sampled uniformly from $\mathcal{A}$, and mutated to generate offspring that are added back to $\mathcal{A}$. To be added, new offspring need to either fill in an empty cell or improve toward the existing elites in their cell. 
The approaches in this work are all based on ME.

\subsubsection{Multi-Objective QD}

QD algorithms conventionally focus on a single fitness objective.
However, this mono-objective approach may be limited in complex real-life scenarios where many goals are involved. 
Multi-Objective QD algorithms (MOQD) \cite{janmohamed2023improving, pierrot2022multi} address this limitation and aim to generate an archive $\mathcal{A}$ with solutions that are not only diverse but also achieve the best possible trade-offs among the objectives to multi-objective problems. 

In general, most MOQD approaches build upon the Multi-Objective MAP-Elites (MOME) algorithm \cite{pierrot2022multi}.
For the most part, MOME follows the ME loop of selection, mutation and addition.
However, MOME differs from ME by allowing more than one solution to be stored in each cell.
In particular, each cell stores a set of solutions that provide the best possible trade-offs across multiple objectives, known as the Pareto Front.
A new solution is added to a cell if it belongs to the Pareto front of the corresponding cell. If the new solution Pareto-dominates solutions that are already in the cell (i.e. scores higher across all objectives) these solutions are removed from the cell.
Recent work~\cite{janmohamed2023improving} has suggested improving MOME by employing crowding-based selection and addition mechanisms which biases the search process towards under-explored regions of the solution space.

\subsection{Solution Reproducibility} \label{sec:reproducibility}

\begin{figure}[h!]
  \centering
  \includegraphics[width=0.95\hsize]{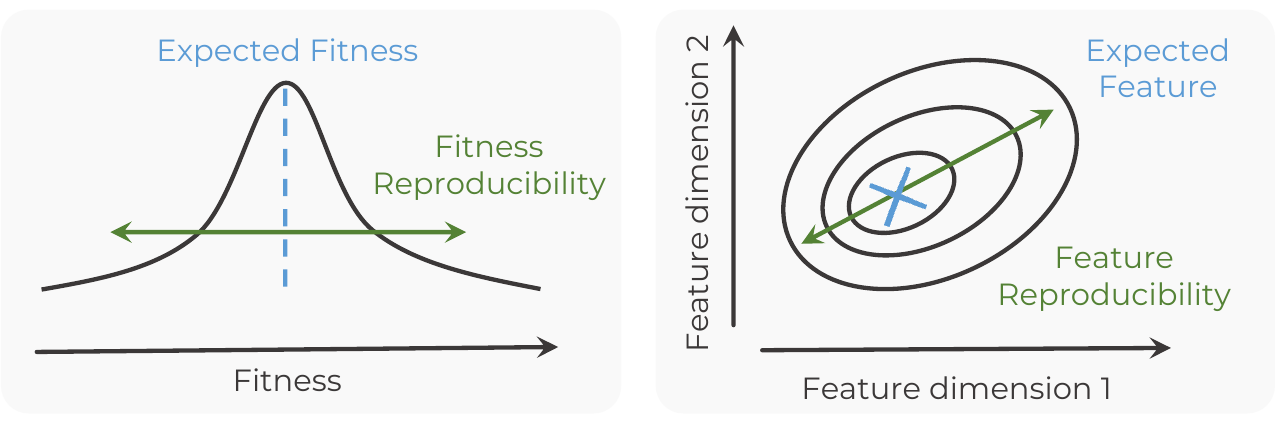}
  \caption{
  \textbf{Solution reproducibility:} Illustration of the expected fitness and feature, and the fitness and feature reproducibilities of a solution.
  }
  \label{fig:uqd}
\end{figure}

The reproducibility of a solution refers to its ability to produce similar results when evaluated multiple times. 
For instance, when controlling a legged robot, a poorly reproducible solution would lead the robot along different trajectories and a highly reproducible solution would strictly follow the same trajectory (Figure~\ref{fig:reproducibility}).
In QD, we can define two types of reproducibility, (see Figure~\ref{fig:uqd}):
\textbf{fitness reproducibility} refers to the statistical dispersion of the fitness distribution, while \textbf{feature reproducibility} refers to the statistical dispersion of the feature distribution. In both cases, a large statistical dispersion means a low reproducibility and vice-versa.
As illustrated in Figure~\ref{fig:reproducibility}, lack of reproducibility can be detrimental and prevent solutions from being effectively deployed.

Feature reproducibility is important for QD because of two main reasons. 
First, when considering the final archive produces by a QD algorithm, downstream solutions are usually sampled from this archive based on their features~\cite{nature, rte}. Feature reproducibility ensures downstream users reliably obtain solutions with the desired features which is critical to the usefulness of the final archive.
Second, when considering the QD optimisation process, solutions with poor feature reproducibility are likely to be added to multiple cells during optimisation, causing the archive to collapse into just a few final solutions with similar features and resulting in significantly reduced diversity in the archive, defeating the purpose of QD.
For these reasons, while some work study both feature and fitness reproducibility~\cite{flageat2023uncertain}, most UQD work consider feature reproducibility only~\cite{grillotti2023don, mace2023quality}. 
This work also focuses on feature reproducibility. From here, we refer to it simply as ``reproducibility", as done in previous work.

\subsubsection{Emergence of reproducibility} 
Many QD algorithms do not account for solution reproducibility, mostly because the environments they consider do not present any uncertainty (i.e. are fully deterministic), so all solutions are entirely reproducible.
Alternatively, some environments present uncertainty that impacts all the solutions in the same way, leading all solutions to have the same reproducibility. For example when the only uncertainty is a fixed-distribution noise on the fitness~\cite{adaptive, flageat2020fast}. 
Thus, reproducibility only becomes critical in environments where each solution can get differently reproducible (referred to as heteroscedastic uncertain~\cite{flageat2024beyond}).
For example in complex robotic control environments where solutions with different speeds might be differently reproducible. 

\subsubsection{Link to robustness}
Reproducibility shares similarities with robustness, a common concept in Robotics and Evolutionary Algorithms~\cite{ea_uncertain}. 
Reproducibility refers to the ability of a solution to replicate consistently its fitness or features when evaluated multiple times within the same uncertain environment, encountered during training. 
Robustness, on the other hand, extends this concept to new variations of the uncertain environment that are only encountered during testing. 
In other words, reproducibility encapsulates whether a solution performs the same for multiple runs of in-distribution environments whereas robustness captures whether a solution can still perform for out-of-distribution ones.
For example, in the robotic example of Figure~\ref{fig:reproducibility}, a reproducible solution would be expected to perform consistently during training and testing within the same environment. In contrast, a robust solution would be expected to perform reliably even in a slightly altered environment, such as one with different friction levels, even if these variations were not encountered during training.
Approaches such as domain randomisation~\cite{tobin2017domain} aim to convert the robustness problem into a reproducibility problem by training on large sets of environment variations, hoping these sets would encompass the possible testing environments.

Jin and Sendhoff~\cite{jin2003trade} describe the trade-off that exists between robustness and fitness for evolutionary algorithms and address it with multi-objective approaches. Their approach is not interested in diversity or behaviour characterisations, however, it is closely related to our work in terms of motivations.

\subsubsection{Quantifying reproducibility and performance} \label{sec:estimators}

We purposefully define reproducibility as a property of solutions, in an estimator-agnostic fashion.
Several estimators have been proposed to quantify the expected performance, such as the mean or median, and to quantify the reproducibility, such as the standard deviation (std) or spread. While a wider overview is given in Appendix B, the algorithms proposed in this work can be used with any estimator. In the following, we use mean for the performance and std for the reproducibility. 

\subsection{Uncertain Quality-Diversity} \label{sec:uqd}

QD algorithms assume the evaluation of solutions returns a reliable estimate of their fitness $f$ and feature $d$, as this is required to compare solutions when adding to the archive $\mathcal{A}$. 
Thus, they do not provide any mechanism to handle cases where these quantities cannot be reliably estimated from a single evaluation. 
In robotic tasks, for example, stochastic dynamics might lead solutions to perform slightly differently from one evaluation to another.
In such environments, QD algorithms tend to keep solutions that obtain ``lucky" evaluations, i.e. outliers whose performance or diversity is overestimated. 
Consider a robotic controller that only walks the robot $1\%$ of the time when conditions are favourable, if it gets lucky during its single evaluation and manages to walk, it would be kept in the QD archive.
These ``lucky" solutions are then returned instead of truly diverse or good-performing ones, thus limiting the effectiveness of QD.
Therefore, Uncertain QD (UQD)~\cite{flageat2023uncertain} designs algorithms for uncertain tasks, where the fitness and feature are no longer fixed values but distributions over possible values: $f \sim \mathcal{D}_f$ and $d \sim \mathcal{D}_d$.

Prior work has identified two main problems for UQD algorithms: the performance estimation problem, that we refer to as \Problemone{} and the reproducibility problem, that we label \Problemtwo{}. Performance estimation (\Problemone{}) refers to the problem of accurately estimating the expected fitness and features of solutions. 
Reproducibility (\Problemtwo{}) refers to the problem of prioritising high-reproducibility solutions as they guarantee repeatable performance (see Section~\ref{sec:reproducibility}).

\subsubsection{Fixed-sampling approaches} \label{sec:fixed_sampling}
These are the most common UQD approaches. They re-evaluate each solution $N$ times, $N$ being a fixed number, to estimate their fitness and feature before addition to $\mathcal{A}$. 
The most common variant, named ME-Sampling in this work, uses the average of the $N$ reevaluations to approximate the fitness and feature of each solution~\cite{hbr, glette_stochastic}.
The ME-Sampling-reproducibility~\cite{flageat2023benchmark} and ME-Low-Spread (ME-LS)~\cite{mace2023quality} variants use the $N$ samples to also estimate solutions reproducibility. In ME-Sampling-reproducibility, a solution is preferred over another if its reproducibility is higher, ignoring its fitness, and in ME-LS, if both its fitness and reproducibility are higher. 
Thus, while ME-Sampling only tackles the performance estimation problem (\Problemone{}), and ME-Sampling-reproducibility only tackles the reproducibility problem (\Problemtwo{}), ME-LS aims to tackle both of them concurrently.
Interestingly, one of the approaches introduced in this work represents a more general case of ME-LS.
Despite being easy to implement, fixed-sampling approaches have a high sampling cost and have been shown to hamper exploration~\cite{flageat2023uncertain, glette_stochastic, ea_uncertain_2}.

\subsubsection{Adaptive-sampling approaches} \label{sec:adaptive_sampling}
Inspired by similar approaches in other fields of optimisation~\cite{ea_adaptive, ea_adaptive_2}, Adaptive-sampling UQD approaches aim to lower the sampling-cost of fixed-sampling approaches. Instead of evaluating every solution multiple times, they only reevaluate promising solutions~\cite{adaptive, flageat2023uncertain}. While they vary in detail, most adaptive-sampling approaches only reevaluate solutions that have a high chance of being added to $\mathcal{A}$, as they are the most promising solutions encountered so far.
In this work, we consider Archive-Sampling (AS)~\cite{flageat2023uncertain}, a simple but effective variant. AS follows the usual ME loop (with $1$ evaluation per offspring) and reevaluates all solutions in $\mathcal{A}$ once per generation.
Most adaptive sampling approaches only tackle the performance estimation problem (\Problemone{}) and ignore the reproducibility problem (\Problemtwo{}). 
While being more efficient than fixed-sampling approaches, they are less frequently applied, mostly due to their additional complexity.

\subsubsection{Other UQD approaches} \label{sec:other_uqd}
Deep-Grid remove the need for sampling by using neighbouring solutions as proxy samples~\cite{flageat2020fast}. 
Alternatively, multiple works have highlighted the benefits of gradient-based mutations for UQD applications~\cite{flageat2023empirical, faldor2023map}. 
Both these approaches do not easily allow reproducibility to be accounted for so we do not consider them in this work.
ARIA~\cite{grillotti2023don} is an optimisation module that improves the performance and reproducibility of the solutions contained in the final collection of any QD algorithm.
While this approach is close to our work, it can be run on top of another UQD algorithm, making it hard to compare. Additionally, ARIA requires an order of magnitude more evaluations than any UQD algorithm. Thus, we do not include it in our results. 
We expand on these choices in Appendix A.

\section{Problem definition} \label{sec:problem}

The objective of UQD optimisation has been formulated in Flageat and Cully~\cite{flageat2023uncertain} as follows: 

\begin{equation} \label{eq:uqd_fit}
\begin{split}
    \max_{\mathcal{A}} \left[ \sum_{e \in \mathcal{A}}{\mathbb{P}_{f_e \sim \mathcal{D}_f} \left[ f_e \right]} \right] 
    \text{  s.t  } \forall e \in \mathcal{A}, \left[ \mathbb{P}_{d_e \sim \mathcal{D}_d} \left[ d_e \right] \right] \in \textrm{cell}_e
\end{split}
\end{equation}

Where $\mathbb{P}$ denotes any estimator of performance (e.g. mean or median; see Section~\ref{sec:estimators}, we note that it does not have to be the same for the fitness and features) and $\textrm{cell}_e$ refers to the feature niche (or cell) of solution $e$ in the QD archive $\mathcal{A}$ (see Section~\ref{sec:qd}). 
In other words, the aim of UQD algorithms is to fill each cell of the archive $\mathcal{A}$ with the solution that has highest chance of belonging to a particular cell and maximum estimated performance, according to the estimator $\mathbb{P}$.

This objective (Equation~\ref{eq:uqd_fit}) does not account for the importance of finding reproducible solutions (we use ``reproducibility" to refer to feature reproducibility as done in previous work; see Section~\ref{sec:reproducibility}). 
Thus, we propose a first modification of Equation~\ref{eq:uqd_fit} that integrates this preference toward reproducible solutions as an additional maximisation constraint:

\begin{equation} \label{eq:uqd_reprod}
\begin{split}
    \max_{\mathcal{A}} \left[ \sum_{e \in \mathcal{A}}{\mathbb{P}_{f_e \sim \mathcal{D}_f} \left[ f_e \right]} \right] 
    &\text{  s.t  } \forall e \in \mathcal{A}, \left[ \mathbb{P}_{d_e \sim \mathcal{D}_d} \left[ d_e \right] \right] \in \textrm{cell}_e \\
    &\text{  s.t  } \forall e \in \mathcal{A}, \max \left[ \mathbb{R}_{d_e \sim \mathcal{D}_d} \left[ d_e \right] \right] 
\end{split}
\end{equation}

Where $\mathbb{R}$ denotes any estimator of the reproducibility (e.g. negative std). 
Even if it is not formulated as such, this new objective (Equation~\ref{eq:uqd_reprod}) corresponds to the one optimised by algorithms proposed in prior work~\cite{grillotti2023don, mace2023quality}. 
However, this paper aim to stress that this objective is, in most cases, unattainable. 
In many UQD applications, ideal solutions which have both high fitness and high reproducibility do not exist. The fitness and reproducibility of solutions might be orthogonal objectives.
For example, in complex robotic control environments, fast-running solutions are intrinsically more brittle than more conservative slower solutions.  
Thus, optimisation in the UQD setting implies making a trade-off between performance and reproducibility, an issue that has been largely underestimated in previous works. 

One major issue that arises from having such a performance-reproducibility trade-off is that 
there is no clear ``best" solution.
The ``better" solution in dependent and must be specified.
Algorithms can only find solutions which achieve a reasonable balance between fitness and reproducibility given some preference which specifies how to prioritise them. 
A user might prefer to have brittle, but super-fast policies, or conversely over-conservative policies, at the cost of their speed. 
In other words, the preference can be seen as part of the optimisation problem definition.
Thus, we propose a second modification of
the UQD objective (Equation~\ref{eq:uqd_fit}) that considers enforcing the performance-reproducibility trade-off preference as an additional constraint:

\begin{equation} \label{eq:uqd_tradeoff}
\begin{split}
    \max_{\mathcal{A}} \left[ \sum_{e \in \mathcal{A}}{\mathbb{P}_{f_e \sim \mathcal{D}_f} \left[ f_e \right]} \right] 
    \text{  s.t  } \forall e \in \mathcal{A}, \left[ \mathbb{P}_{d_e \sim \mathcal{D}_d} \left[ d_e \right] \right] \in \textrm{cell}_e \\ 
    \text{  s.t  } \forall e \in \mathcal{A}, \textrm{preference} \left[ \mathbb{P}_{f_e \sim \mathcal{D}_f}  \left[ f_e \right], \mathbb{R}_{d_e \sim \mathcal{D}_d} \left[ d_e\right] \right] \\
\end{split}
\end{equation}

Where preference refers to any mapping given by a final user that specifies which solution should be preferred over another given their respective fitness and reproducibility.
Section~\ref{sec:uqd} defines the two problems commonly identified in UQD. In this work, we propose a third one: the \textbf{\problem{}}, that corresponds to  enforcing preferences over the trade-off between performance and reproducibility. 
Equation~\ref{eq:uqd_tradeoff} encompasses these three problems.


\section{Method} \label{sec:method}

In this section, we introduce one possible method to formulate preferences in the \problem{}: the \parametrisation{}. 
We also propose $5$ new UQD algorithms based on this parametrisation. 
Again, we use ``reproducibility" to refer to feature reproducibility (see Section~\ref{sec:reproducibility}).

\subsection{\parametrisation{}} \label{sec:parametrisation}

\begin{figure}[h!]
  \centering
  \includegraphics[width=0.95\hsize]{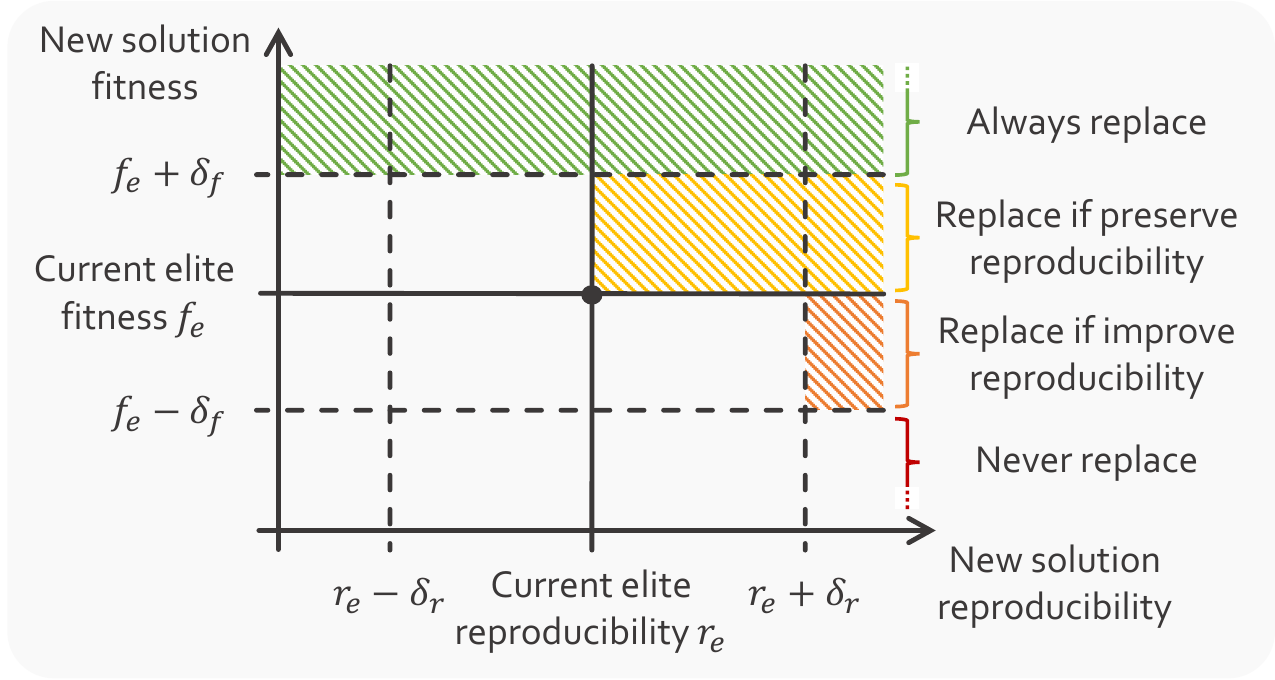}
  \caption{
  \textbf{\parametrisation{}}: we consider a new solution compared to an existing elite $e$. The x-axis represents the reproducibility of this new solution, and the y-axis its fitness. We place the reproducibility $r_e$ and fitness $f_e$ of the elite $e$ on these axes. Coloured areas indicate values for which the new solution replaces the elite $e$.
  }
  \label{fig:parametrisation}
\end{figure}

We formulate preferences over performance-reproducibility trade-offs using an introduced \parametrisation{}.
The main assumption of the \parametrisation{} is that fitness is always the primary objective of UQD algorithms, and we can use fitness as a ``reference" objective.
The \parametrisation{} consists of two parameters $\paramr{}$ and $\paramf{}$, which can be defined and interpreted as follows: \textit{``an increase of $\paramr{}$ in reproducibility compensates for a loss of $\paramf$ in fitness"}.
We visualise the \parametrisation{} in Figure~\ref{fig:parametrisation}. 
We note that this parametrisation is agnostic to the performance and reproducibility estimators (see Section~\ref{sec:estimators}).

In many cases, it will be possible to infer $\paramr{}$ and $\paramf{}$ beforehand from the task or from the targeted application, so algorithms can use these values during optimisation. We refer to this setting as \textit{a-priori}~\cite{marler2010weighted}. 
Alternatively, it is also possible to have an \textit{a-posteriori} setting \cite{marler2010weighted} where preference parameters are unavailable before optimisation.
Algorithms in this setting will maintain solutions with different trade-off values to allow users to decide on one after optimisation.

\subsection{A-priori Approaches: Weighted Sum}\label{sec:weighted_sum}

The \parametrisation{} allows us to define a first approach that adjusts the fitness value to account for reproducibility using a weighted sum. 
Inspired by Marler and Arora~\cite{marler2010weighted}, we propose the following adjusted fitness, where $f$ is the fitness estimate and $r$ the reproducibility estimate:

\begin{equation}\label{eq:weighted_sum}
    \tilde{f} = f + \frac{\paramf{} + \rho}{\paramr{} + \rho} r
\end{equation}

Where $\rho$ is an arbitrarily small number that has two functions: (1) it ensures that $\tilde{f}$ is defined when $\paramr{} = 0$ and (2) it ensures that, between two solutions of equal fitness, the more reproducible one will always be preferred, even when $\paramf{} = 0$. 
The main advantage of this approach is its simplicity as it requires marginal change to existing UQD algorithms. We propose two variants of it:

\subsubsection{Fixed-sampling weighed-sum}
Integrating the weighted-sum in fixed-sampling approaches (Section~\ref{sec:fixed_sampling}) is quite straightforward: each solution is sampled $N$ times to estimate its fitness and reproducibility, and the algorithm optimises for the weighted sum from Equation~\ref{eq:weighted_sum}. 

\subsubsection{Adaptive-sampling weighted-sum}
The integration in adaptive-sampling approaches (Section~\ref{sec:adaptive_sampling}) is less straightforward as it raises the question of non-stationarity of the estimated-performance of a solution.
As solutions are reevaluated, their estimated performance changes with time and early good solutions might be lost due to unfavourable initial evaluations.
Previous adaptive-sampling approaches have overcome this issue by adding a depth to the ME grid, allowing multiple solutions to be stored in a cell and thus giving further opportunities for underestimated solutions that might prove promising after further evaluations. 
We also use cell depth in our approaches to tackle this issue. 

\subsection{A-priori Approaches: Delta Comparison} \label{sec:delta}

The \parametrisation{} also allows the implementation of a delta-comparison-based archive addition. 
When choosing if a new solution $i$ should replace an elite $e$, instead of relying on fitness-only, we use the following criteria (also in Figure~\ref{fig:parametrisation}):

\begin{equation} \label{eq:delta}
\begin{split}
	\text{New\_elite}(i, e) = 
        \begin{cases} 
            i &\text{ if } f_i \geq f_{e} + \paramf{}\\
            i &\text{ if } f_i \geq f_{e} \text{ and } r_i \geq r_{e}\\
            i &\text{ if } f_i \geq f_{e} - \paramf{} \text{ and } r_i \geq r_{e} + \paramr{}\\
            e &\text{ otherwise }
        \end{cases}
\end{split}
\end{equation}

Where $f_i$ and $f_e$ are the fitness estimates of solution $i$ and elite $e$, and $r_i$ and $r_e$ are their reproducibility estimates. 
This criteria shares similarities with epsilon dominance~\cite{laumanns2002combining}.

\subsubsection{Fixed-sampling delta-comparison}
fixed-sampling approaches (Section~\ref{sec:fixed_sampling}) can integrate delta-comparison by sampling solutions $N$ times to estimate their fitness and reproducibility, and add to the archive based on Equation~\ref{eq:delta}.

Interestingly, the previously introduced ME-LS approach~\cite{mace2023quality} (see Section~\ref{sec:fixed_sampling}) is a particular case of fixed-sampling delta-comparison with $\paramf{} = 0$ and $\paramr{} = 0$.

\subsubsection{Adaptive-sampling delta-comparison}
The integration in adaptive-sampling approaches (Section~\ref{sec:adaptive_sampling}) raises the question of non-stationarity already mentioned in the weighted-sum case, and hence requires the use of a depth. 
However, a depth of $d$ would usually keep the $d$ best-performing solutions encountered so far; while such ranking is easy to obtain for single-score comparison, it becomes more complex when using delta-comparison. 
To overcome this, we propose the following: when a new solution is considered for addition within a cell, it is compared to the elites one by one in descending order until it dominates one of them and replaces it. The replaced elite is then similarly compared to the less-performing elites of the cell.


\subsection{A-posteriori Approaches: Multi-Objective QD}\label{sec:posteriori}

It may not always be possible to specify a preference parameter \textit{a-priori}. 
For example, the user may have limited domain knowledge of the task or problem, making it difficult to suggest a suitable preference or alternatively, they may simply wish to seek solutions which present a variety of trade-offs.
In such scenarios, we propose using multi-objective QD (MOQD) algorithms~\cite{pierrot2022multi} in order to find a range of solutions that present trade-offs of fitness and reproducibility, for each cell of the archive.
This way, the user can first examine the range of possible trade-offs that are possible and then select their preference \textit{a-posteriori}.
In this work, we use the MOME algorithm~\cite{pierrot2022multi} with crowding-based exploration~\cite{janmohamed2023improving}, which we term MOME-X \cite{janmohamed2024multi}.
We note that a similar approach was used as a baseline in Grillotti et al~\cite{grillotti2023don} (see Section~\ref{sec:other_uqd}). However, this approach was tackling the problem of reproducibility maximisation instead of trade-off enforcement (see Section~\ref{sec:problem}) and only one final best solution was kept per cell. Additionally, it was based on MOME instead of MOME-X and showed less promising results than ours.

We note that maintaining a range of solutions that present different trade-offs is a key advantage of \textit{a-posteriori} approaches, as it affords the end user flexibility to change between preferences in different cells, or at different times.
Additionally, if a preference parameter becomes available, then we can sample one solution per cell by ``projecting" the final grid into a standard ME grid. 
Given some values $\paramf{}$ and $\paramr{}$, we project the MOME archive by choosing the solution in each cell of the grid which has the maximum weighted sum objective as defined in Equation~\ref{eq:weighted_sum}.

\section{Main experimental results} \label{sec:robotics}

\subsection{Experimental Setup}

\subsubsection{Algorithms}
We report the results for the $5$ new algorithms described in Section~\ref{sec:method}: 
\begin{itemize}
    \item MOME-X: \textit{a-posteriori} fixed-sampling.
    \item ME-Weighted: \textit{a-priori} weighted-sum fixed-sampling.
    \item ME-Delta: \textit{a-priori} delta-comparison fixed-sampling.
    \item AS-Weighted: \textit{a-priori} weighted-sum adaptive-sampling.
    \item AS-Delta: \textit{a-priori} delta-comparison adaptive-sampling.
\end{itemize}

Of these methods, all of the fixed-sampling approaches use $32$ samples and all the adaptive-sampling approaches are based on Archive-Sampling (AS, see Section~\ref{sec:uqd}) with $2$ initial samples.
We compare to the baselines from Section~\ref{sec:uqd}:
\begin{itemize}
    \item Vanilla-ME.
    \item ME-Sampling with $32$ samples.
    \item ME-Sampling-Reproducibility with $32$ samples.
    \item ME-LS with $32$ samples.
    \item Vanilla-Archive-Sampling (Vanilla-AS).
\end{itemize}

All our implementations use the QDax library~\cite{chalumeau2024qdax, lim2022accelerated}.
To facilitate later works, we open-source our comparison code at \url{https://github.com/adaptive-intelligent-robotics/Uncertain_Quality_Diversity}.

\begin{table}[ht!]
\centering
\footnotesize

  \caption{Task suite considered in this work.}
  \begin{tabular}{ c | c c c }

    & \textsc{Hexapod} &
    \textsc{Walker} & \textsc{Ant} \\

    & \includegraphics[width = 0.08\textwidth]{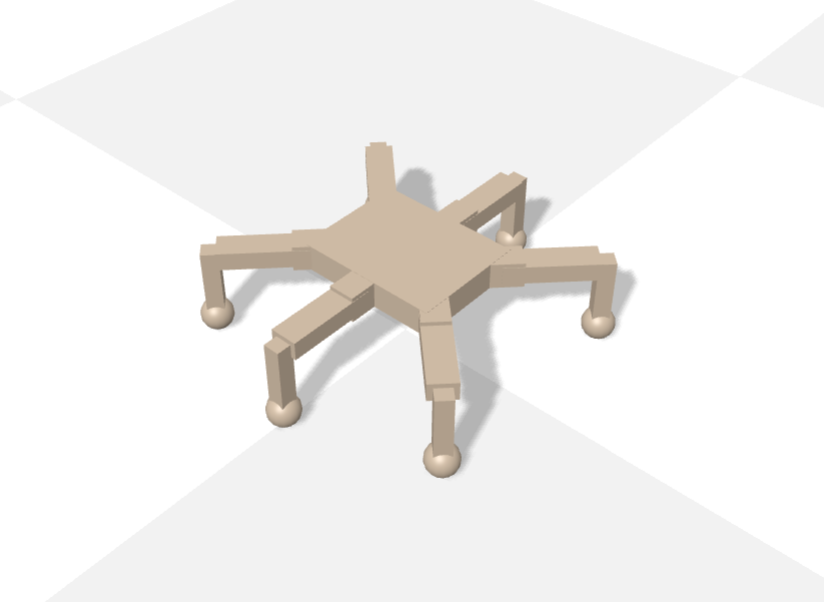} 
    & \includegraphics[width = 0.08\textwidth]{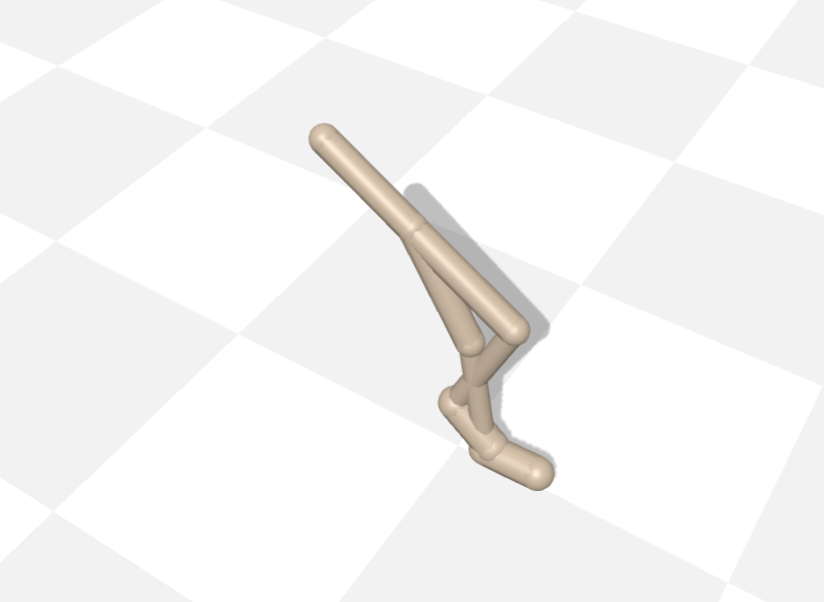} 
    & \includegraphics[width = 0.08\textwidth]{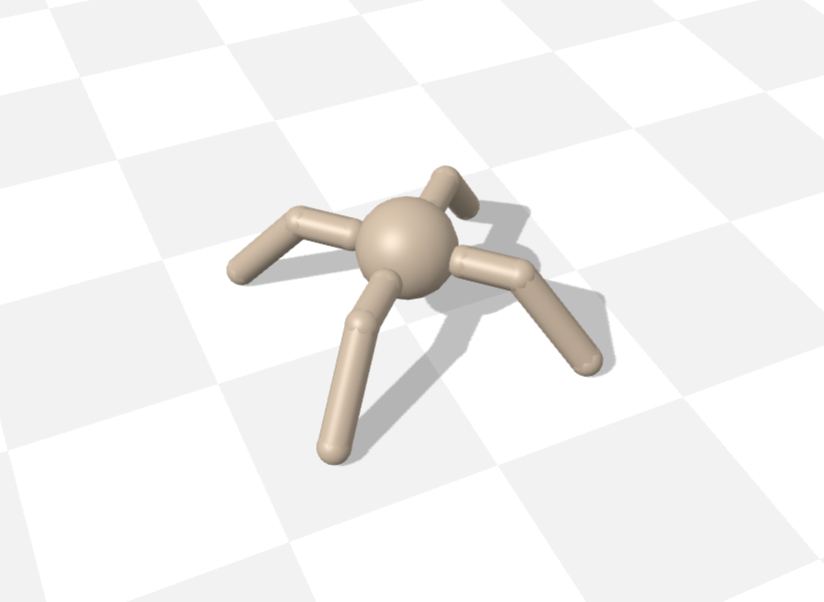} \\

    

    
    \midrule
    
    \textsc{Fitness} 
    & \makecell{Orientation \\ error}
    & \makecell{Speed bonus, \\ energy penalty, \\  survival bonus}
    & \makecell{Energy penalty, \\ survival bonus} \\
    
    \midrule
    
    \textsc{Feature} 
    & \makecell{Final position}
    & \makecell{Feet contact}
    & \makecell{Final position} \\

    
    \midrule
    
    \textsc{Uncertainty} 
    & \makecell{Noise on fitness \\ and feature.}
    & \multicolumn{2}{c}{\makecell{Noise on initial joint \\ positions and velocities.}} \\
    
    \midrule
    
    \textsc{[$\paramf{}$, $\paramr{}$]} 
    & $[140, 0.14]$
    & $[260, 0.04]$
    & $[220, 6.0]$ \\

    
  \end{tabular}
  \label{tab:tasks}
\end{table}

\subsubsection{Tasks}
We consider the robotic control tasks in Table~\ref{tab:tasks}, used in previous work~\cite{flageat2023uncertain, flageat2023empirical, mace2023quality, grillotti2023don, flageat2022benchmarking} (further information provided in Appendix C).
For $\paramf{}$ and $\paramr{}$, we ran some initial experiments to determine the range of fitness and reproducibility values for each task, then, we picked $10\%$ of these ranges for unidirectional tasks and $20\%$ for omnidirectional tasks as values for $\paramf{}$ and $\paramr{}$. 
We emphasise that we did not optimise these values but picked them arbitrarily to simulate realistic user's preferences. 

\subsubsection{Metrics}
We compare performance using the \textbf{Corrected QD-Score} and the \textbf{Reproducibility-Score}. 
To calculate the Corrected QD-Score, we approximate the ``real" performance of each solution as the median of $512$ reevaluations (shown to be a reasonable number in Flageat et al.~\cite{flageat2023uncertain}) and add all of them back into an empty archive known as the Corrected Archive. The Corrected QD-Score is the QD-Score~\cite{pugh} of this Corrected archive.
We also approximate the reproducibility of solutions using these $512$ reevaluations. 
To do so, we
first get the descriptor variance of the $512$ reevaluations of
each solution, normalised within each cell using the maximum
observed-variance. This normalisation accounts for the difference in descriptor variance across the descriptor space. The
Reproducibility-Score is then computed as the sum over the
archive of 1 - normalised variance, to avoid penalising approaches that find more solutions.

All algorithms were run for $10$ seeds for each task (giving a total of $330$ runs), and we report $p$-values from a Wilcoxon signed-rank test with a Holm-Bonferroni correction.

\subsubsection{Sampling-size comparison}
QD algorithms are usually run with a fixed batch-size, which for vanilla QD refers both to the number of evaluations per generation and to the number of offspring per generation. However, these two quantities are no longer the same in the UQD setting where most algorithms use complex sampling strategies. 
Flageat et al.~\cite{flageat2023uncertain} proposed instead to run UQD algorithms with a fixed maximum budget of evaluations per generation, referred to as the sampling-size. 
For example, with sampling-size $16384$, ME generates $16384$ offspring per generation, but ME-Sampling with $32$ samples generates $16384 / 32 = 512$ offspring per generation. AS uses $2048$ samples to reevaluate the content of the archive (i.e. the number of cells), thus it generates $16384 - 2048 = 14336$ offspring per generation. In the following we use a sampling-size of $16384$ for all our comparisons.

\subsection{Results}

\begin{figure*}[t!]
  \centering
  \includegraphics[width=\hsize]{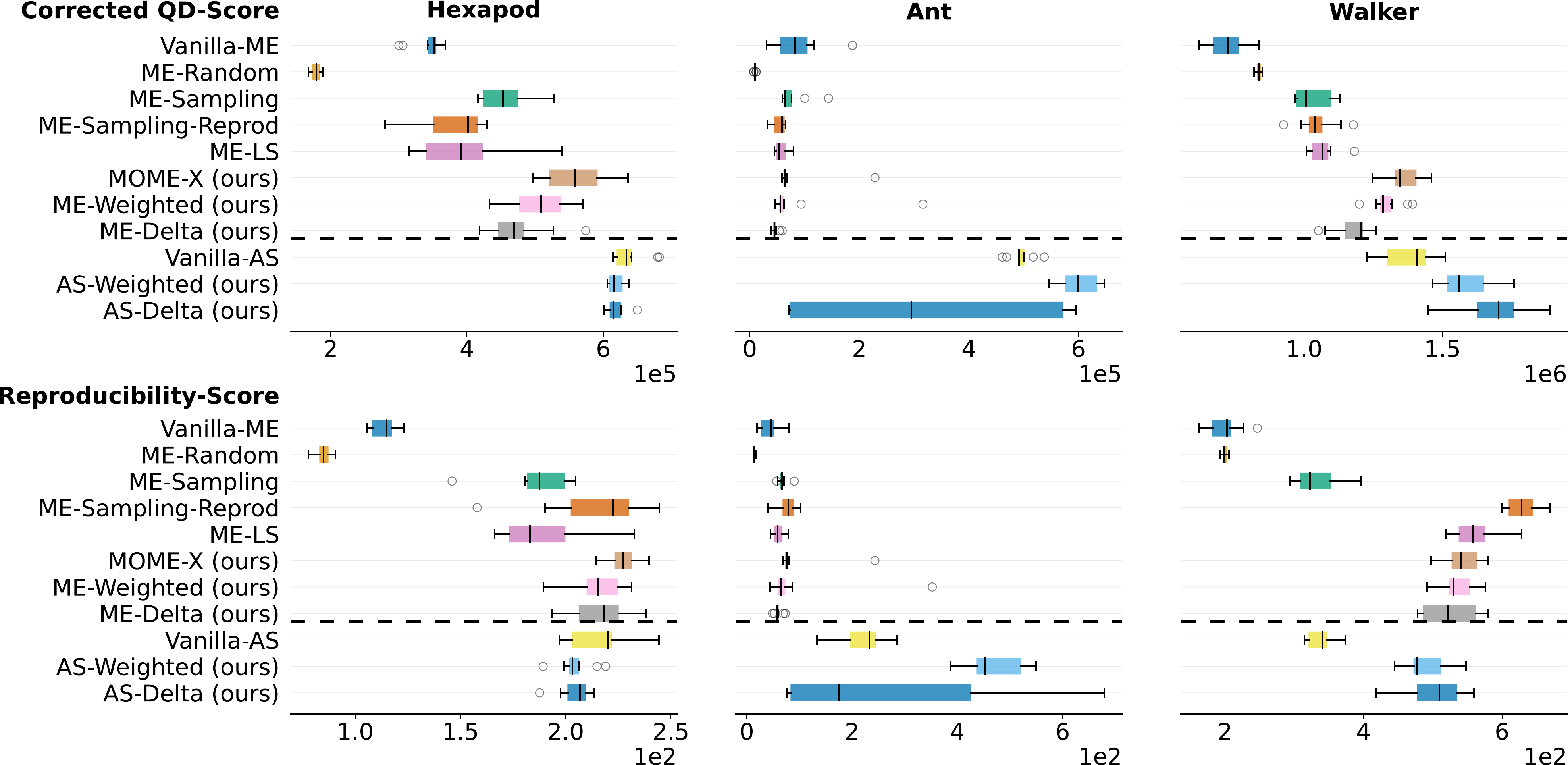}
  \caption{
    \textbf{Robotic tasks results:} (top) Corrected QD-Score, displaying the quality and diversity of the final archive, and (bottom) Reproducibility-Score, quantifying the reproducibility of the solutions in the final archive. 
    For both metrics, higher score is better. 
    The vertical lines show the median across $10$ replications, the boxes the quartiles, the whiskers $1.5$ times the interquartile range, and the dots represent outliers. 
    Each plot is split (by horizontal lines) into two parts: fixed-sampling approaches (both baselines and proposed approaches) and adaptive-sampling (both baselines and proposed approaches).
  }
  \label{fig:robotic}
\end{figure*}

\begin{figure}[t!]
  \centering
  \includegraphics[width=\hsize]{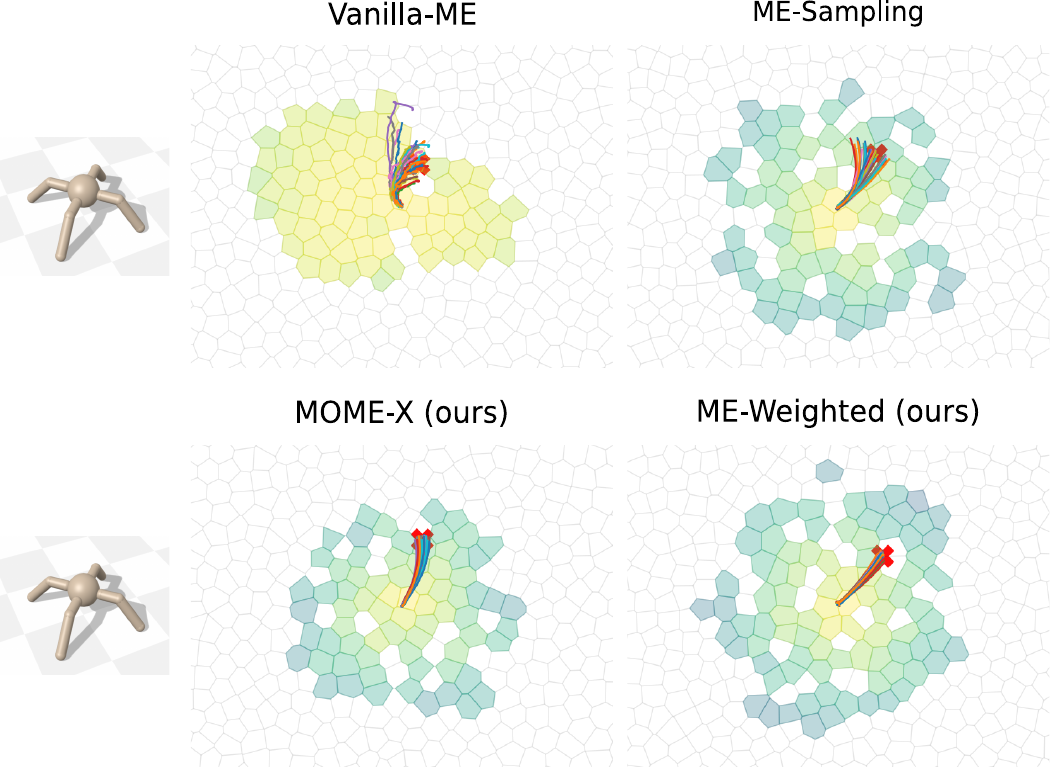}
  \caption{
  \textbf{Example reproducibilities:} trajectories obtained by the same policy replicated $32$ times in the Ant environment, displaying the importance of reproducibility. We randomly sample $1$ of the $10$ seeds of each algorithm and $1$ feature value, we replicate $32$ times the corresponding solutions and plot the resulting $32$ trajectories. We also display the full archive as a background for each algorithm, and the target feature as a red cross. The larger the spread of the trajectories the lower the reproducibility of the solution.
  }
  \label{fig:robotic_reproducibilities}
\end{figure}

We display the results in Figure~\ref{fig:robotic}.
For each task, we distinguish two categories of approaches, separated on the plot by horizontal lines: fixed-sampling approaches (both baselines and proposed fixed-sampling approaches), and adaptive-sampling approaches (both baselines and proposed fixed-sampling approaches). 
We also display in Figure~\ref{fig:robotic_reproducibilities} some example trajectories, illustrating the reproducibilities of solutions found by different approaches. 

\subsubsection{Main takeaway}
Our key result is that our proposed approaches outperform all UQD baselines in diversity and quality of the final archive on Walker (Corrected QD-Score; $p<0.005$) and Hexapod ($p<0.1$, $p<0.01$ for MOME-X).
This is a striking result as our approaches were designed to solve another problem (i.e. enforcing preferences over trade-off values), and not to outperform existing baselines.
Moreover, we emphasise that $\paramf{}$ and $\paramr{}$ were chosen arbitrarily for those tasks and it is likely that these values could be optimised to achieve even better performance.
These results indicate that simply accounting for the performance-reproducibility trade-off leads to outperforming approaches that ignore it.

\subsubsection{Detailed comparison with UQD baselines}
As mentioned earlier, our $3$ fixed-sampling approaches outperform previous fixed-sampling approaches in Corrected QD-Score on Walker and Hexapod. 
The only exception is the Ant task, known to be difficult to solve with fixed-sampling~\cite{flageat2023uncertain, flageat2023empirical}. 
We hypothesise that this is because the task is deceptive: policies that walk in a reliable manner are rare and are surrounded in the search space by more brittle solutions that tend to fall easily. 
Thus, reaching areas of the search space where solutions walk consistently requires using brittle solutions as stepping stones. 
Due to their fixed number of reevaluations, fixed-sampling approaches systematically reject these intermediate stepping stones. They get quickly stuck with small archives of a few solutions and therefore there is almost no renewal of the pool of parents.
On the contrary, adaptive-sampling approaches keep these stepping stone solutions and are therefore able to overcome this deceptive area. 
Our two new adaptive-sampling approaches outperform previous adaptive-sampling approaches on Walker ($p<0.001$) and perform similarly on Hexapod. 
However, while our weighted adaptive-sampling approach also outperform baselines on Ant ($p<0.001$), the delta-comparison variant appears to obtain bimodal results and not systematically solve the task. We hypothesise this might be due to the deceptive structure of the task, which only some replications manage to overcome.

In term of Reproducibility-Score, our approaches outperform Vanilla-ME, ME-Random and ME-Sampling ($p<0.01$). As could be expected, they perform similarly to ME-LS, which is a particular case of the proposed ME-Delta approach (see Section~\ref{sec:delta}). 
ME-Sampling-Reproducibility, which optimises only for reproducibility, proves better on Walker ($p<0.001$). This is likely due to the value chosen for $\paramf{}$ and $\paramr{}$ that do not allow our approaches to keep the most reproducible solutions of all.
Our $2$ AS approaches also outperform Vanilla-AS on Walker ($p<0.001$) and on Ant for AS-Weighted only ($p<0.001$), but perform similarly on Hexapod. As the Hexapod is controlled using an open-loop controller, the reproducibility of solutions is probably capped and it seems likely that all our approaches have reach a plateau that Vanilla-AS was already reaching.

\subsubsection{A-priori versus a-posteriori}
Across all tasks, the a-posteriori fixed-sampling approach (MOME-X) outperforms the two a-priori approaches in Corrected QD-Score ($p<0.01$ for Walker and $p<0.1$ otherwise).
We emphasise that these metrics were calculated on the projected archives (see Section~\ref{sec:posteriori}) and that alternative projection strategies or alternative $\paramf{}$ and $\paramr{}$ choices could achieve even higher performance on these metrics, for no additional cost.

\section{Benchmark tasks results} \label{sec:benchmark}

Flageat et al.~\cite{flageat2023benchmark} introduced more targeted UQD Benchmark tasks to analyse and estimate the performance of UQD algorithms. These tasks are split into three categories: Performance-estimation, Reproducibility-maximisation and Realistic tasks. 
The two first categories specifically target the two main problems of the UQD setting (\Problemone{} and \Problemtwo{} ; see Section~\ref{sec:uqd}). 
In the following, we introduce and propose a set of tasks for the third UQD problem introduced in this work: the \problem{} (see Section~\ref{sec:problem}).
We report the performance of our algorithms on this extended UQD Benchmark suite.

\subsection{Results on Existing UQD Benchmarks} \label{sec:reproducibility_benchmark}

\begin{figure}[t!]
  \centering
  \includegraphics[width=\hsize]{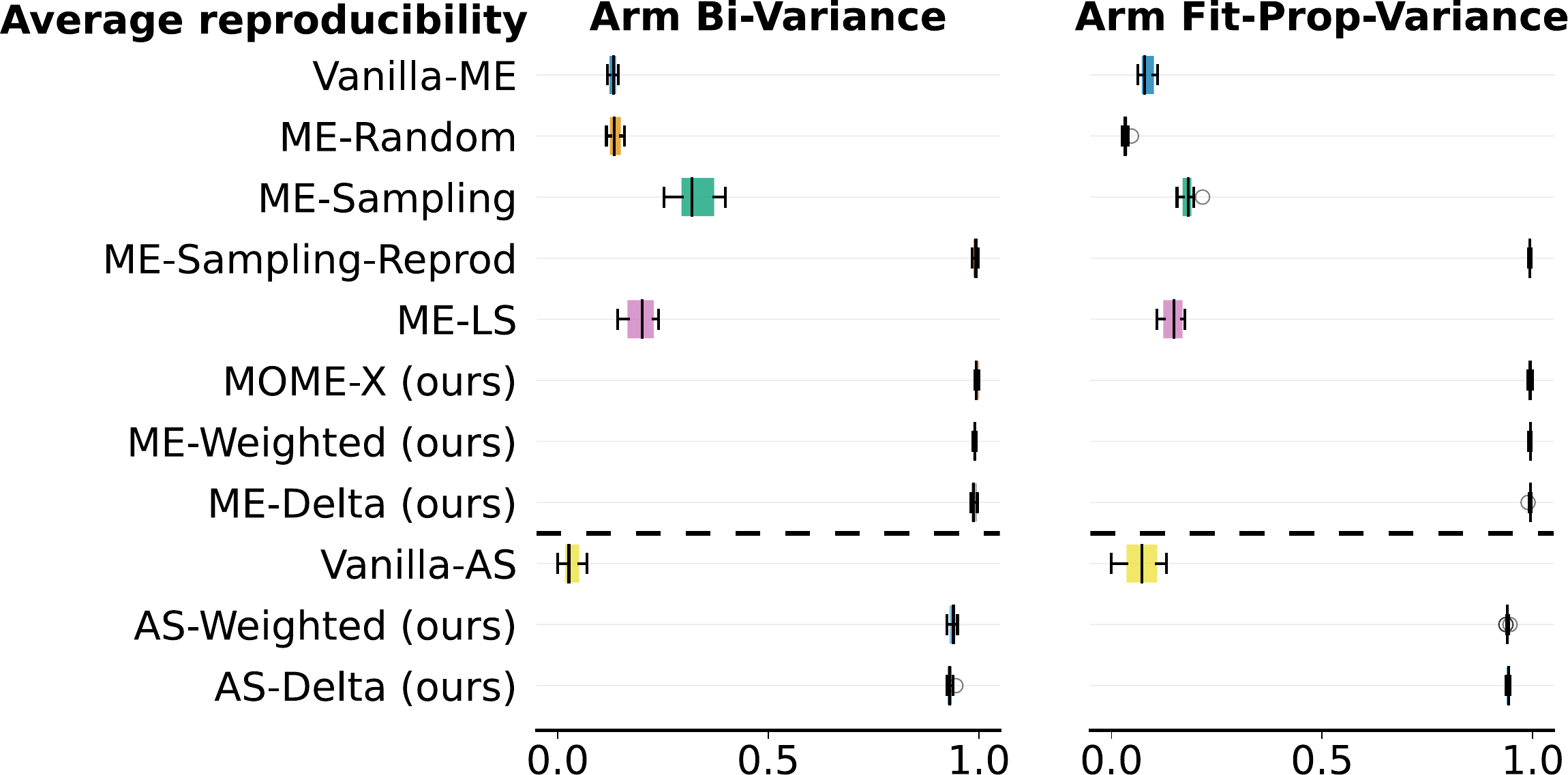}
  \caption{
    \textbf{Reproducibility-maximisation benchmark results:} in these tasks, the fitness is always $0$, so we only display the reproducibility averaged over the archive. For this metric, a higher score is better, and the maximum value is $1$.  
    The vertical lines show the median across $10$ replications.
  }
  \label{fig:reproducibility_benchmark}
\end{figure}

We first provide the results on the Reproducibility-maximisation benchmark tasks~\cite{flageat2023benchmark} in Figure~\ref{fig:reproducibility_benchmark}.
In these two tasks, all solutions are allocated a fitness value of $0$ so there is no performance-reproducibility trade-off and the only goal is to find the most reproducible solutions in each cell, making ME-Sampling-Reproducibility a strong baseline.
We display the \textbf{Average Reproducibility}, averaged over the archive.

The $3$ fixed-sampling approaches proposed in this work find solutions as reproducible as ME-Sampling-Reproducibility according to the Average Reproducibility, thus outperforming every other baseline ($p<0.001$) and solving the task.
However, the $2$ adaptive-sampling approaches find solutions that are slightly less reproducible. This highlights the main limitation of adaptive-sampling approaches: their tendency to use a lower number of samples can lead to important estimation errors that mislead the algorithm. 
This limitation is not manifest in complex robotics tasks as in Section~\ref{sec:robotics} where their sample-efficiency and ability to overcome deceptive landscapes allow them to outperform other approaches.
However, this limitation becomes noticeable in simpler tasks such as this one which are not deceptive and where sample-efficiency is not critical.

\subsection{Extending UQD Benchmarks} \label{sec:tasks}

\begin{figure}[t!]
  \centering
  \includegraphics[width=0.92\hsize]{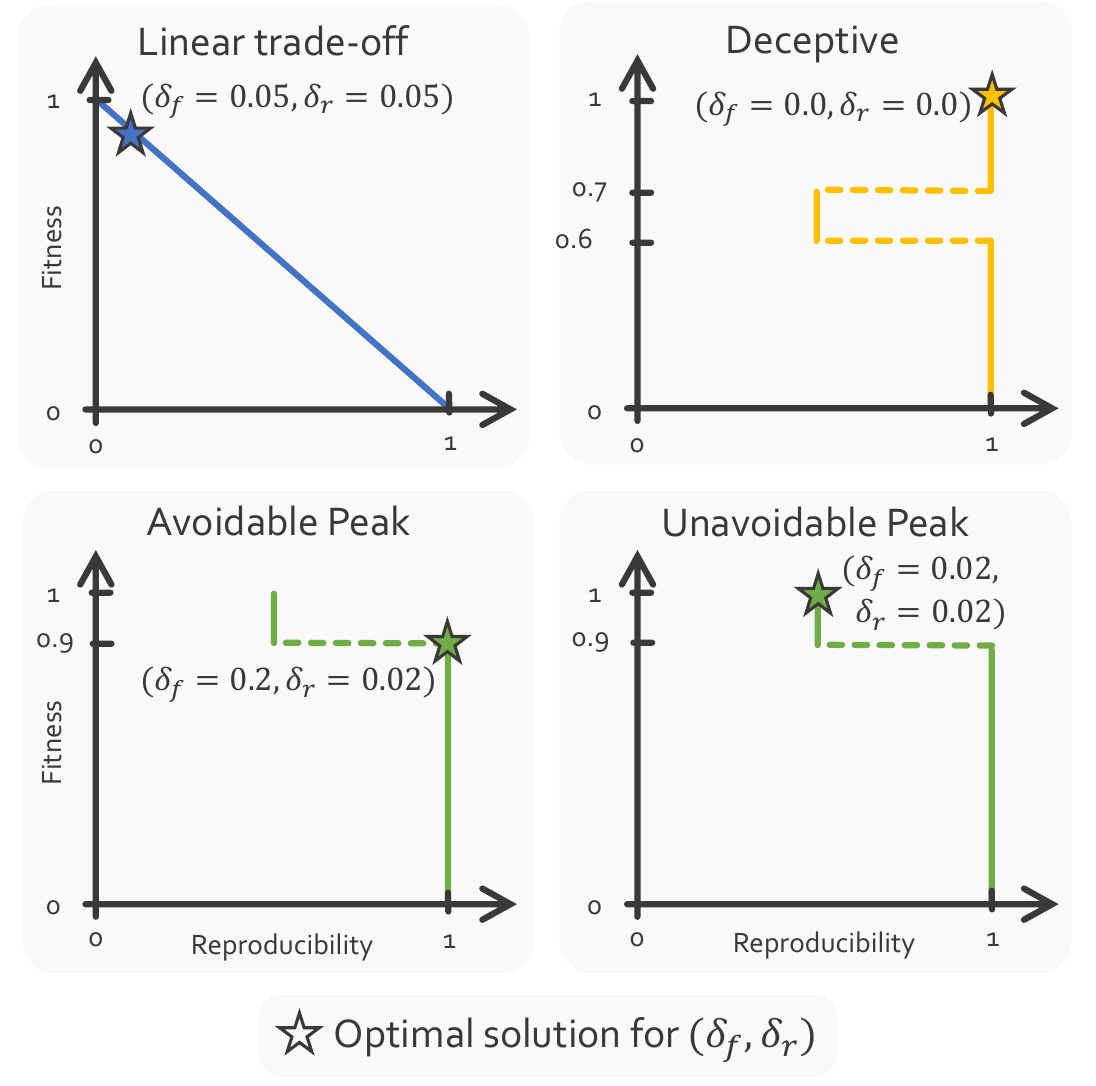}
  \caption{
  \textbf{Proposed trade-off benchmark tasks:} representation for each new task of the performance-reproducibility profile, which gives the relationship between reproducibility and fitness.
  In these tasks, $(\paramf{}, \paramr{})$ are part of the task definition, and we represent the corresponding optimal solution (star).
  }
  \label{fig:tasks}
\end{figure}

We propose extending the UQD Benchmark tasks~\cite{flageat2023benchmark} with four new benchmark tasks targeting the \problem{}.
Each task has a specific relationship between reproducibility and fitness, that we refer to as performance-reproducibility profiles, illustrated in Figure~\ref{fig:tasks}. 
In these tasks, $(\paramf{}, \paramr{})$ are given as part of the task definition, thus there is one optimal solution in each cell, and algorithms are expected to find them to solve the task.

\begin{itemize}
    \item \textbf{Linear trade-off:} an increase of $\delta$ in fitness leads to a loss of $\delta$ in reproducibility and vice-versa. 
    We set $(\paramf{} = 0.05, \paramr{} = 0.05)$ and expect algorithms to find optimal solutions that achieve this trade-off.
    \item \textbf{Deceptive:} aims to spot algorithms that get stuck in their exploration due to solutions with intermediate fitness but low reproducibility. We set $(\paramf{} = 0.0, \paramr{} = 0.0)$.
    \item \textbf{Avoidable and Unavoidable Peaks:} these $2$ tasks present a low-reproducibility peak for the highest fitness values. 
    For the Avoidable Peak, we set $(\paramf{} = 0.2, \paramr{} = 0.02)$, so $\paramf{}$ is bigger than the peak and algorithms should select the high-reproducibility solution just before the peak. 
    For the Unavoidable Peak, we set $(\paramf{} = 0.02, \paramr{} = 0.02)$, so $\paramf{}$ is smaller than the peak and algorithms should select the high-fitness solution despite their low reproducibility.
\end{itemize}

We define these tasks in the ``Direct Mapping" environment, where the $3$-dimensional genotype directly encodes the fitness and the two features dimensions. 
The uncertainty comes from a Gaussian noise added to the features. The variance of the Gaussian depends on the fitness following the performance-reproducibility profile of the task. For example, for the Linear trade-off task, the variance of the Gaussian noise on the features is directly proportional to the value of the fitness.

We choose this simple ``Direct Mapping" environment because it guarantees that any cell in the feature space can contain any value of fitness and reproducibility. 
Thus, the optimal value of fitness to enforce the desired trade-off is the same for all the cells.
This allows us to display archive-wide metrics that capture the algorithm's performance. In comparison, in more complex tasks, only part of the fitness and reproducibility values would be attainable in each cell and capturing the performance would require per-cell metrics.

\subsection{Results on Extended UQD Benchmarks}

\begin{figure}[t!]
  \centering
  \includegraphics[width=0.98\hsize]{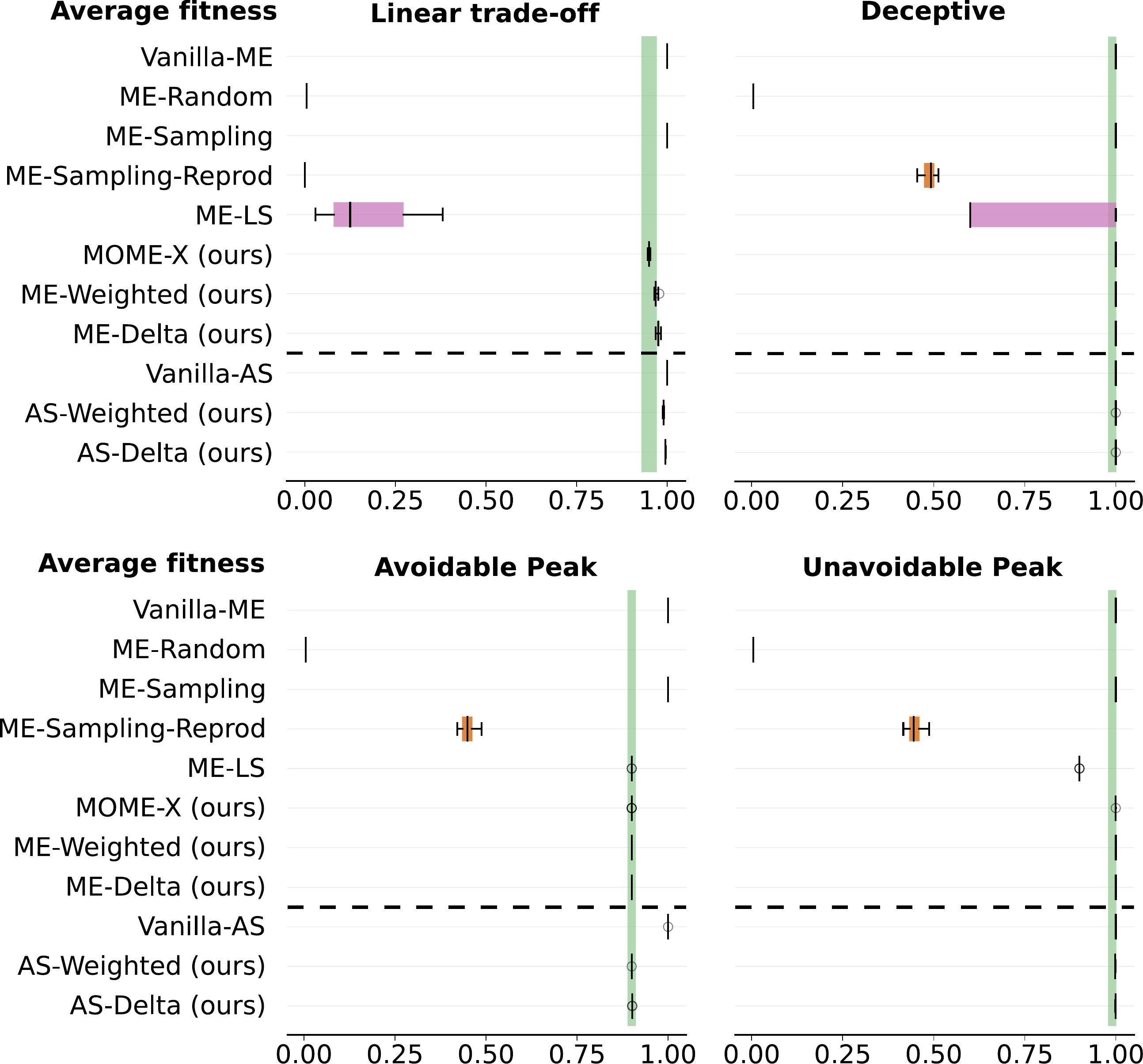}
  \caption{
  \textbf{Trade-off benchmark results:} we report the fitness averaged over the archive corresponding to the performance-reproducibility profiles in Figure~\ref{fig:tasks}. The green-coloured areas indicate fitness values that algorithms are expected to converge to. 
  The vertical lines show the median across $10$ replications.
  }
  \label{fig:tradeoff_benchmark}
\end{figure}

We provide the results on the Trade-Off benchmark tasks in Figure~\ref{fig:tradeoff_benchmark}.
We report the \textbf{Average Fitness} which relates to reproducibility following the profiles in Figure~\ref{fig:tasks}. 

\subsubsection{Linear} 
In this first task, the $3$ fixed-sampling approaches converge to the required trade-off value and solve the task.
However, the $2$ adaptive-sampling approaches do not manage to converge to the exact required value. This highlights again the tendency of adaptive-sampling approaches to perform estimation error as mentioned in Section~\ref{sec:reproducibility_benchmark}.

\subsubsection{Deceptive} 
All approaches manage to solve the Deceptive case except ME-Sampling-Reproducibility and ME-LS. While this is expected for ME-Sampling-Reproducibility, this highlights an important limitation of the ME-LS approach: its strict conservation of both fitness and reproducibility prevents it from systematically overcoming such deceptive traps. It seems reasonable to assume that more complex search space might have inherent deceptive traps, and this characteristic might explain the lower performance of ME-LS in Section~\ref{sec:robotics}.

\subsubsection{Avoidable and Unavoidable Peak} 
These $2$ tasks best highlight the performance of our proposed approaches. 
When given $\delta$-parameters that require to avoid the low reproducibility peak, all our approaches do so and find the maximum fitness before the peak. When given $\delta$-parameters that make the peak unavoidable, they all converge to the maximum fitness.
In comparison, approaches that do not account for the trade-off stop systematically at the same trade-off value, oblivious to the $\delta$-parameters, either before the peak (ME-LS and ME-Sampling-Reproducibility), or after the peak (all others). 

\section{Conclusion and Discussion}

In this work, we introduce the \problem{} to the field of Uncertain QD.
We propose the \parametrisation{} as a method for specifying preferences over trade-offs and develop $5$ approaches that use this parametrisation to solve this new problem: $2$ a-priori fixed-sampling, $2$ a-priori adaptive-sampling and $1$ a-posteriori fixed-sampling approaches. 
We experimentally demonstrate that our approaches successfully implement preferences. 
Importantly, we also show that by simply accounting for the performance-reproducibility trade-off, our approaches outperform existing QD methods in uncertain domains, while they were not designed to do so.
This suggests that considering the performance-reproducibility trade-off unlocks important stepping stones that are usually missed when only performance is optimised.

A limitation of our work is that it only focuses on feature reproducibility and does not account for fitness reproducibility, similar to previous work. While we believe all our approaches can be extended to take fitness reproducibility into account, either via considering the three-objective case or via a mixture of fitness and feature reproducibilities, we leave this dimension for future work. 
We also leave open the definition of a-posteriori adaptive-sampling algorithms, which could be done by reevaluating the MOME archive periodically as done in Archive-Sampling in this work. 
We hope this work raises awareness of the importance of taking into account the performance-reproducibility trade-off in Uncertain QD optimisation and opens the door to new algorithms that more effectively handled this Uncertain QD setting. 




\bibliographystyle{IEEEtran}
\bibliography{biblio}

\newpage

\begin{IEEEbiography}[{\includegraphics[width=1in,height=1.25in,clip,keepaspectratio]{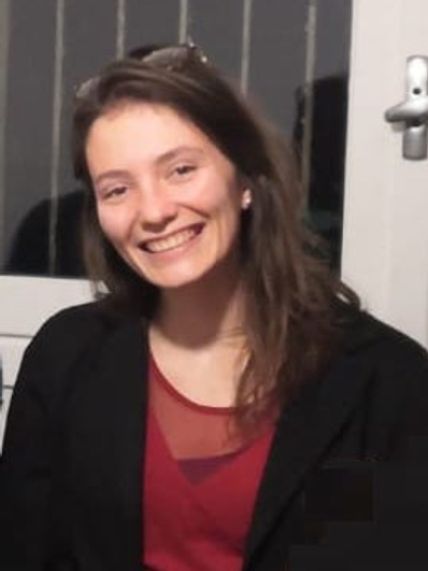}}]{Manon Flageat}
is a PhD student in the Adaptive and Intelligent Robotics Laboratory, Department of Computing, Imperial College London.
Her research focuses on Quality-Diversity algorithms, in particular applied to uncertain environments, as well as Deep Reinforcement Learning and synergies between these two types of learning algorithms.
Before, she received an MSc degree in Biomedical Engineering (specialism: Neurotechnology) from Imperial College London (United Kingdom) in 2019, and an engineering degree from the École Nationale Supérieure des Mines de Saint-Étienne, France, in 2019. 
She is currently also a Teaching Scholar in the Department of Computing, Imperial College London. 
\end{IEEEbiography}

\begin{IEEEbiography}[{\includegraphics[width=1in,height=1.25in,clip,keepaspectratio]{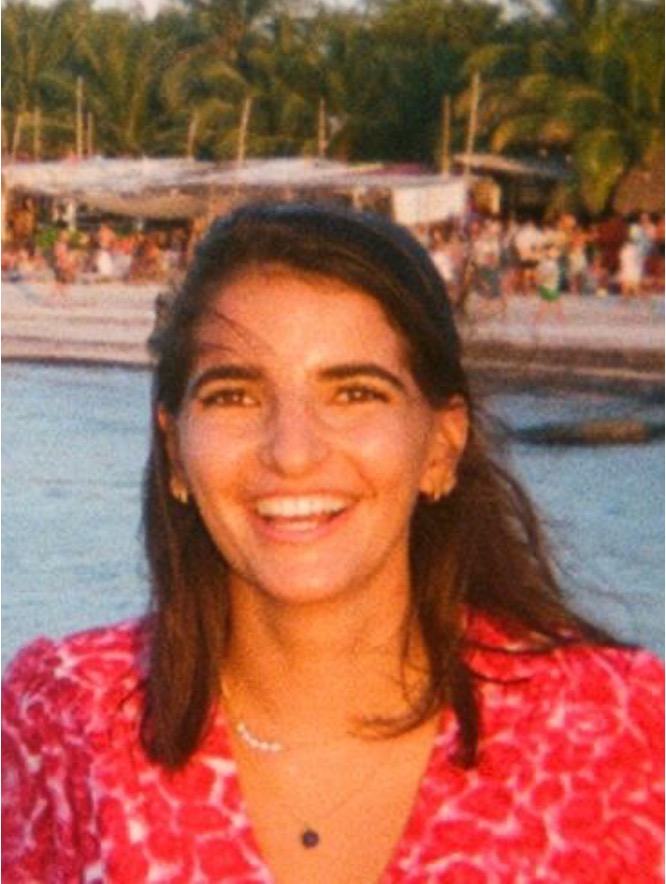}}]{Hannah Janmohamed}
received an MSc Computing (Artificial Intelligence and Machine Learning) degree from Imperial College London (United Kingdom) in 2022, and a BSc Mathematics degree from Durham University in 2019. 
She is currently an InstaDeep scholar and PhD student in machine learning for robotics at the Adaptive and Intelligent Robotics Laboratory, Department of Computing, Imperial College London.
Her research focuses on Multi-Objective Quality-Diversity algorithms, in applications ranging from robotics to material design. Hannah also currently serves as the President of Women and Non-Binary Individuals in Computing Society, Imperial College London. 
\end{IEEEbiography}

\begin{IEEEbiography}[{\includegraphics[width=1in,height=1.25in,clip,keepaspectratio]{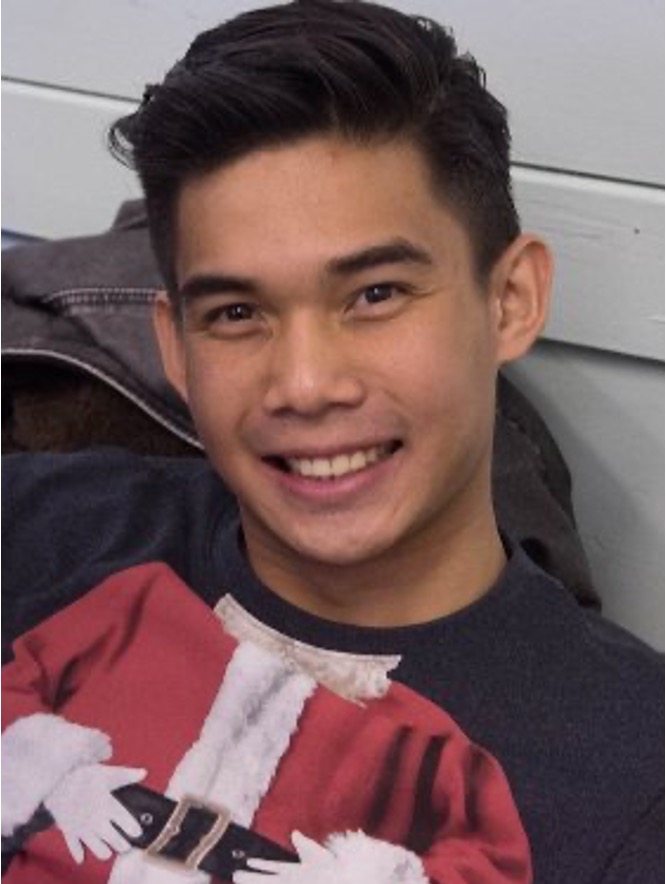}}]{Bryan Lim}
is a Senior AI Research Scientist at Autodesk. He focuses on open-ended learning systems which can continuously generate a diversity of interesting problems and corresponding novel solutions, with the potential to lead to increasingly intelligent, creative and general-purpose AI systems. To enable such open-ended systems, he works on increasing the efficiency and scalability of Quality-Diversity algorithms, which encourages novelty and diversity to enable more creative search processes. Bryan’s work is at the intersection of reinforcement learning, robotics and evolutionary computation and he has co-authored papers in venues such as ICLR, ICRA, GECCO, ALIFE and TMLR. Bryan obtained his PhD from Imperial College London in 2024. Before that, he completed his MEng year abroad at MIT and has an undergraduate degree in Mechanical Engineering from Imperial College London.
\end{IEEEbiography}

\begin{IEEEbiography}[{\includegraphics[width=1in,height=1.25in,clip,keepaspectratio]{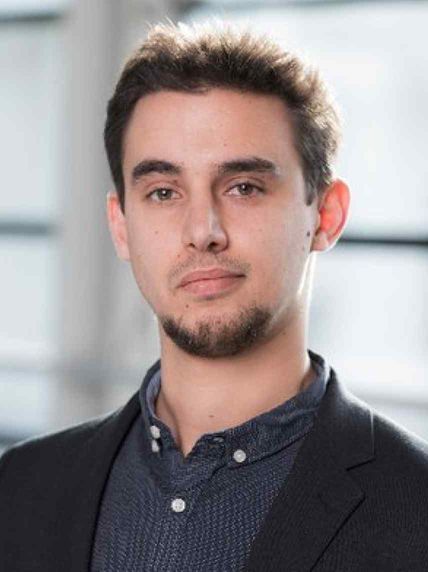}}]{Antoine Cully}
is Reader (Associate Professor) at Imperial College London (United Kingdom) and the director of the Adaptive and Intelligent Robotics Lab. His research is at the intersection between artificial intelligence and robotics. He applies machine learning approaches, like evolutionary algorithms, on robots to increase their versatility and their adaptation capabilities. In particular, he has recently developed Quality-Diversity optimization algorithms to enable robots to autonomously learn large behavioural repertoires. For instance, this approach enabled legged robots to autonomously learn how to walk in every direction or to adapt to damage situations.
Antoine Cully received the M.Sc. and the Ph.D. degrees in robotics and artificial intelligence from the Sorbonne Université in Paris, France, in 2012 and 2015, respectively, and the engineering degree from the School of Engineering Polytech’Sorbonne, in 2012. His Ph.D. dissertation has received three Best-Thesis awards. He has published several journal papers in prestigious journals including Nature, IEEE Transaction in Evolutionary Computation, and the International Journal of Robotics Research. His work was featured on the cover of Nature (Cully et al., 2015), received the "Outstanding Paper of 2015" award from the Society for Artificial Life (2016), the French "La Recherche" award (2016), and four Best-Paper awards from GECCO (2021, 2022, 2023, 2024).
\end{IEEEbiography}

\newpage

\section{Other UQD approaches} 
\label{app:other_uqd}

This Section complements Section II.C.3 and details our motivation to exclude other UQD algorithms from our study.

\subsection{Implicit-sampling approaches} 

Implicit-sampling approaches~\cite{ea_uncertain, ea_uncertain_2} remove the need for sampling by using neighbouring solutions as samples-proxy. To the best of our knowledge, the only UQD implicit-sampling approaches are Deep-Grid~\cite{flageat2020fast} and Deep-Grid-sampling~\cite{flageat2023uncertain}.
These approaches prove promising in simple UQD environments, however, they tend to be limited when the genotype-feature mapping is complex~\cite{flageat2023empirical}. They also tend to find solutions less reproducible than other approaches~\cite{flageat2023uncertain}.
Additionally, existing work in implicit-sampling has not yet proposed any mechanism to estimate reproducibility using the evaluations of neighbouring solutions. As reproducibility is at the core of our study, this limitation prevents us from developing a-priori or a-posteriori implicit-sampling approaches. 
For these reasons, we do not use them in this work.

\subsection{Gradient-augmented QD approaches}

Previous works~\cite{flageat2023empirical, faldor2023map} have also highlighted the benefits of gradient-augmented QD approaches in UQD domains thanks to their modelling of the environment. 
These approaches are not explicitly designed for UQD settings but only prove beneficial as a side effect of their optimisation strategies. 
Additionally, they rely on additional assumptions as they only apply to the Markov Decision Process setting. Thus, as done in previous work~\cite{flageat2023uncertain}, we do not consider them in this work.

\subsection{ARIA} 
ARIA~\cite{grillotti2023don} is an optimisation module that improves the performance and reproducibility of the solutions contained in the final collection returned by any QD algorithm.
This approach can be run on top of another UQD algorithm, and thus improve the results of any UQD algorithm. This first dimension makes the comparison difficult as it is unclear which part of the results can be inputted to ARIA itself and it would require running it with all our baselines and approaches. 
Additionally, ARIA requires an order of magnitude more evaluations than any UQD algorithms. 
Finally, ARIA assumes information about the structure of the ME grid to quantify reproducibility (i.e. the probability of belonging to a cell), which none of the approaches in this paper consider.
Thus, we choose to exclude this algorithm from our comparison.

\begin{figure*}[h!]
  \centering
  \includegraphics[width=\hsize]{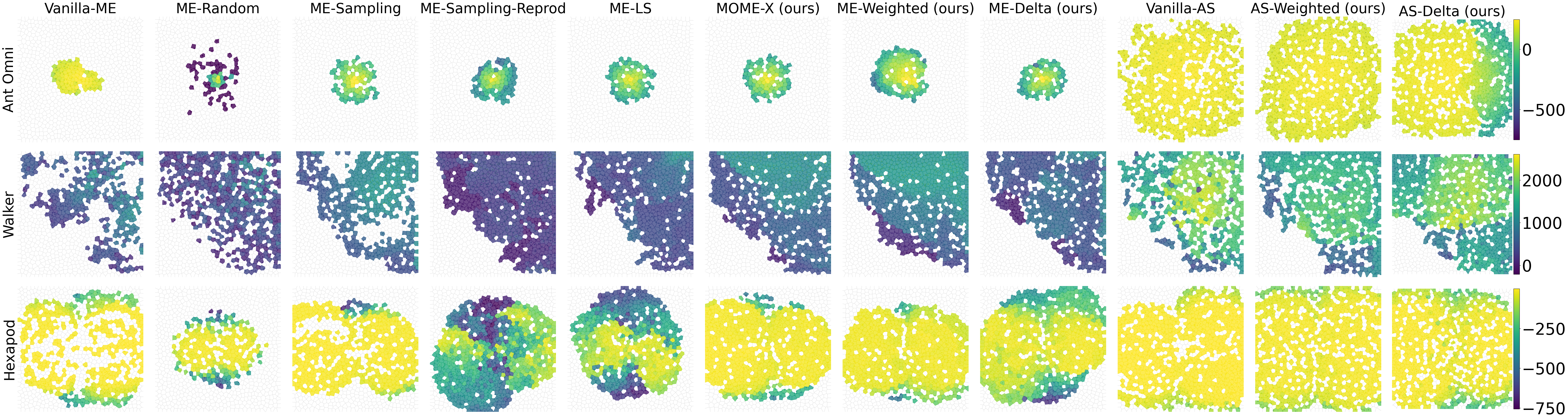}
  \caption{
    \textbf{Robotic control tasks - Corrected archives:}
    we report the archive for $1$  randomly sampled seed out of the $10$ seeds. 
    For each subplot, the $x$ and $y$ axes correspond to the first and second feature dimensions respectively. The colour corresponds to fitness (the brighter the better; see the colour bar on the right).
  }
  \label{fig:archives_reeval_robotic}
\end{figure*}

\begin{figure*}[h!]
  \centering
  \includegraphics[width=\hsize]{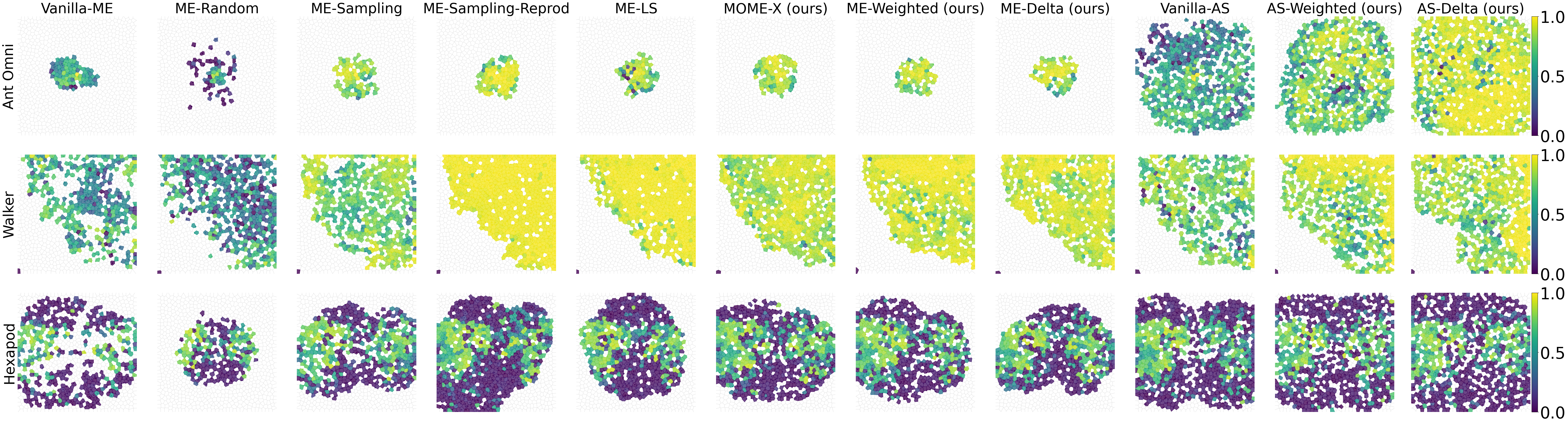}
  \caption{
    \textbf{Robotic control tasks - Reproducibility archives:} 
    we report the archive for $1$  randomly sampled seed out of the $10$ seeds. 
    For each subplot, the $x$ and $y$ axes correspond to the first and second feature dimensions respectively. The colour corresponds to reproducibility (the brighter the better; see the colour bar on the right).
  }
  \label{fig:archives_reproducibility_robotic}
\end{figure*}

\section{Performance and Reproducibility Estimators} \label{app:estimators}

As mentioned in Section II.B.3, performance and reproducibility are purposefully defined in an estimator-agnostic manner. Here we aim to list the estimators that have been used in other algorithms.  
Previous work has used the following estimators for reproducibility:
(a) standard deviation (std)~\cite{flageat2023uncertain, flageat2023benchmark, flageat2024beyond, mace2023quality}, (b) median absolute deviation (mad)~\cite{flageat2024beyond}, (c) inter-quartile range (iqr)~\cite{flageat2024beyond}, (d) probability to belong to the cell (tailored for the ME-based algorithms)~\cite{grillotti2023don}. 
Similarly, estimators for fitness and features include (a) mean~\cite{flageat2020fast, adaptive, flageat2024beyond, grillotti2023don}, (b) median~\cite{flageat2023uncertain, flageat2023benchmark}, (c) closest to median~\cite{flageat2023uncertain}, (d) mode~\cite{mace2023quality}. 
We re-iterate that all the approaches proposed in this work can be used with any of those estimators.

\begin{table}[h!]
\centering
\small

  \caption{Task suite considered in this work.}
  \begin{tabular}{ c | c c c }

    & \textsc{Hexapod} &
    \textsc{Walker} & \textsc{Ant} \\

    & \includegraphics[width = 0.08\textwidth]{figures/hexa.png} 
    & \includegraphics[width = 0.08\textwidth]{figures/walker.png} 
    & \includegraphics[width = 0.08\textwidth]{figures/ant.png} \\
    \addlinespace[0.05cm]

    \addlinespace[0.05cm]
    \midrule
    \addlinespace[0.05cm]
    
    \addlinespace[0.05cm]
    \textsc{Control} 
    & \makecell{Periodic \\ functions}  
    & \makecell{Neural \\ network}
    & \makecell{Neural \\ network} \\
    
    \textsc{Dims} & $36$ & $198$ & $296$ \\


    \addlinespace[0.05cm]
    
    \addlinespace[0.05cm]
    \midrule
    \addlinespace[0.05cm]
    
    \addlinespace[0.05cm]
    \textsc{Fitness} 
    & \makecell{Orientation \\ error}
    & \makecell{Speed bonus, \\ energy penalty, \\  survival bonus}
    & \makecell{Energy penalty, \\ survival bonus} \\
    \addlinespace[0.05cm]
    
    \addlinespace[0.05cm]
    \midrule
    \addlinespace[0.05cm]
    
    \addlinespace[0.05cm]
    \textsc{Feature} 
    & \makecell{Final position}
    & \makecell{Feet contact}
    & \makecell{Final position} \\

    \textsc{Dims} & $2$ & $2$ & $2$ \\

    \addlinespace[0.05cm]
    
    \addlinespace[0.05cm]
    \midrule
    \addlinespace[0.05cm]
    
    \addlinespace[0.05cm]
    \textsc{\makecell{Uncer- \\ tainty}} 
    & \makecell{Noise on \\ fitness and \\ features}
    & \multicolumn{2}{c}{\makecell{Noise on initial joint \\ positions and velocities.}} \\
    \addlinespace[0.05cm]
    
    \addlinespace[0.05cm]
    \midrule
    \addlinespace[0.05cm]
    
    \addlinespace[0.05cm]
    \textsc{$\paramf{}$} 
    & $140$
    & $260$
    & $220$ \\

    \textsc{$\paramr{}$} 
    & $0.14$
    & $0.04$
    & $6.0$ \\
    
  \end{tabular}
  \label{tab:complete_tasks}
\end{table}

\section{Robotics Experimental Setup}\label{app:setup}

This section gives in Table~\ref{tab:complete_tasks} the complete detail of the robotic experimental setup used in this work.

\section{Archives} \label{app:archives}

\begin{figure*}[h!]
  \centering
  \includegraphics[width=\hsize]{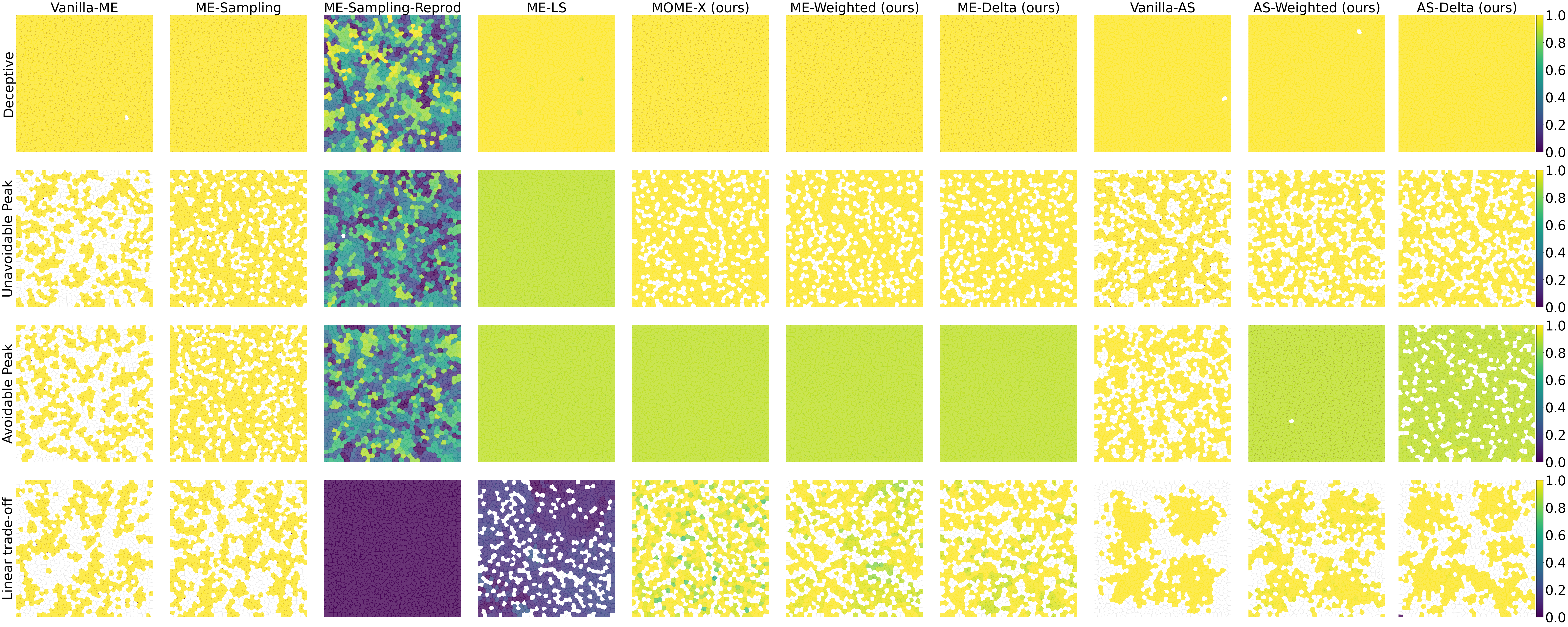}
  \caption{
    \textbf{Trade-off benchmark - Corrected archives:}
    we report the archive for $1$  randomly sampled seed out of the $10$ seeds. 
    For each subplot, the $x$ and $y$ axes correspond to the first and second feature dimensions respectively. The colour corresponds to fitness (the brighter the better; see the colour bar on the right). The reproducibility can be inferred from the fitness value based on the fitness-reproducibility profile of the task given in Figure~\ref{fig:tasks}.
  }
  \label{fig:archives_benchmark_tradeoff}
\end{figure*}

\begin{figure*}[h!]
  \centering
  \includegraphics[width=\hsize]{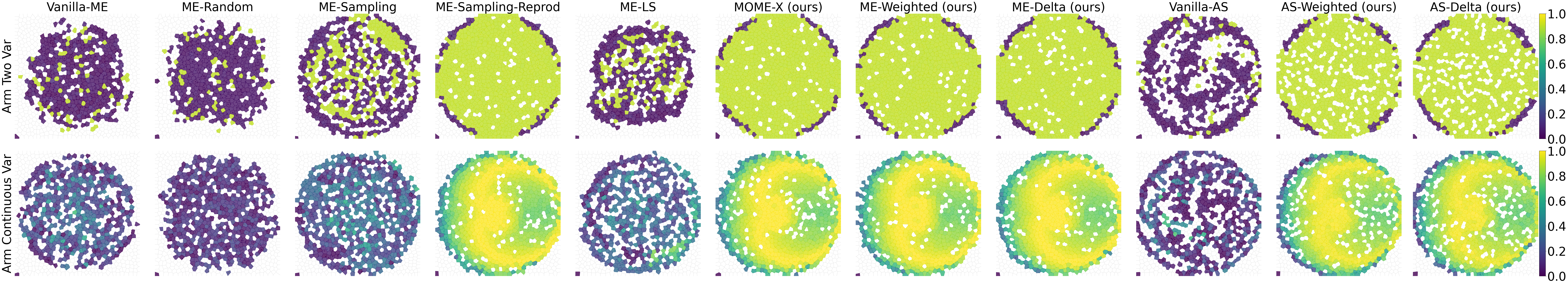}
  \caption{
    \textbf{Reproducibility-maximisation benchmark - Reproducibility archives:} 
    we report the archive for $1$  randomly sampled seed out of the $10$ seeds. 
    For each subplot, the $x$ and $y$ axes correspond to the first and second feature dimensions respectively. The colour corresponds to reproducibility (the brighter the better; see the colour bar on the right). 
  }
  \label{fig:archives_benchmark_reproducibility_archives}
\end{figure*}

We give in Figure~\ref{fig:archives_reeval_robotic} and~\ref{fig:archives_reproducibility_robotic} the Corrected archives and the Reproducibility archives respectively, for the robotics tasks considered in Section V.B.
We also give in Figure~\ref{fig:archives_benchmark_tradeoff} the Corrected archives of the new benchmark tasks proposed in Section VI.C.
For these tasks, the reproducibility can be inferred from the fitness value based on the fitness-reproducibility profile of the task given in Figure~\ref{fig:tasks} so we do not provide the reproducibility archives. 
In Figure~\ref{fig:archives_benchmark_reproducibility_archives}, we report the Corrected archives for the UQD reproducibility benchmark tasks considered in Section VI.A. 
The fitness in this task is set to $0$ so we do not report any fitness archives.

\section{Note on diversity-reproducibility trade-off}

This paper defined the performance-reproducibility trade-off for the UQD setting. 
In this section, we wish to argue that it is not possible to define similarly a diversity-reproducibility trade-off in UQD. 
The common formulation and objective of QD algorithms is to maximise the coverage of the feature space (see Section II.A).
This objective prevents "choosing" to fill one feature neighbourhood over another: every neighbourhood of the space has to contain one solution. 
Thus, while there may be regions of the feature space with inherently lower reproducibility, it is not possible to "choose" the reproducible feature neighbourhood over the non-reproducible ones. Both have to be filled and thus there is no diversity-reproducibility trade-off in UQD.

\end{document}